\pdfoutput=1

\documentclass[11pt]{article}
\usepackage[utf8]{inputenc}
\usepackage[preprint]{acl}
\usepackage{times}
\usepackage{latexsym}
\usepackage{amssymb}
\usepackage{graphicx}
\usepackage{verbatim}
\usepackage{newunicodechar}
\usepackage{lmodern}
\usepackage{comment}
\usepackage{url}
\usepackage{lettrine}
\usepackage{hyperref}
\usepackage{footnote}
\usepackage{manyfoot}
\usepackage{textgreek}
\usepackage{fancyvrb}
\usepackage{fvextra}
\usepackage{fancyvrb}
\usepackage{enumitem}
\usepackage{hyperref}
\usepackage{multirow}
\usepackage{rotating}
\usepackage{booktabs}

\usepackage{times}
\usepackage{algorithm}
\usepackage{colortbl} 
\usepackage{amsmath}
\usepackage{algorithmic}
\usepackage{subcaption}
\usepackage{todonotes}
\usepackage{amssymb}
\usepackage{textcomp}

\newunicodechar{−}{-}
\DefineVerbatimEnvironment{CustomVerbatim}{Verbatim}{fontsize=\small, frame=single, breaklines=true, breaksymbolleft={}}

\usepackage{microtype}

\usepackage{inconsolata}
\usepackage{booktabs}
\usepackage{multirow}
\usepackage{graphicx}

\usepackage{colortbl}

\title{\benchmark: Evaluating Grounded Multi-Turn Chart Editing in Multimodal Language Models}

\newcommand{\benchmark}{ChartEditBench}

\author{Manav Nitin Kapadnis\footnotemark[1] ~~ Lawanya Baghel\footnotemark[1] ~~ Atharva Naik\footnotemark[1] ~~ Carolyn Ros\'e \\
  Language Technologies Institute \\
  Carnegie Mellon University \\
  \texttt{ \{iammanavk, lawanyabaghel\}@gmail.com, \{arnaik, cprose\}@cs.cmu.edu}}

\begin{document}
\maketitle
\begin{abstract}
While Multimodal Large Language Models (MLLMs) perform strongly on single-turn chart generation, their ability to support real-world exploratory data analysis remains underexplored. In practice, users iteratively refine visualizations through multi-turn interactions that require maintaining common ground, tracking prior edits, and adapting to evolving preferences.
We introduce \benchmark, a benchmark for incremental, visually grounded chart editing via code, comprising 5,000 difficulty-controlled modification chains and a rigorously human-verified subset. Unlike prior one-shot benchmarks, \benchmark evaluates sustained, context-aware editing.
We further propose a robust evaluation framework that mitigates limitations of LLM-as-a-Judge metrics by integrating execution-based fidelity checks, pixel-level visual similarity, and logical code verification.
Experiments with state-of-the-art MLLMs reveal substantial degradation in multi-turn settings due to error accumulation and breakdowns in shared context, with strong performance on stylistic edits but frequent execution failures on data-centric transformations. \benchmark, establishes a challenging testbed for grounded, intent-aware multimodal programming.
\end{abstract}

\renewcommand{\thefootnote}{\fnsymbol{footnote}}
\footnotetext[1]{Equal contribution.}

\section{Introduction}
\label{Chapter 1}
Recent advances in multimodal large language models (MLLMs) have driven rapid progress in \emph{chart generation}: given a natural language description or an input image, models produce plotting code or synthesized visualizations, supported by large synthetic corpora and dedicated benchmarks \cite{chart-r1, masry2022chartqa, masry-etal-2025-chartqapro}. 
. Although LLMs often saturate these single-turn generation and question answering benchmarks \cite{masry-etal-2025-chartqapro}, this framing overlooks a core aspect of real-world visualization workflows.
In practice, developers, analysts, and technical writers rarely create a visualization in a single step; instead, they iteratively refine existing specifications, adjust encodings, add statistical overlays, and reformat layouts in response to evolving requirements and feedback. 
Consequently, editing an existing chart specification represents a natural and high-value capability for MLLM-based assistants, yet remains comparatively underexplored.

To investigate whether MLLMs can understand and maintain common ground with complex user intents over multiple turns, we introduce \benchmark, a synthetic data generation framework for evaluating chart editing via code modification. 
Specifically, \benchmark assesses a model’s ability to update an existing visualization in response to either natural language instructions or a target image specifying the desired changes. 
We further construct multi-turn interaction sequences in both settings, enabling evaluation of whether models can track incremental, semantically meaningful updates across both the textual and visual domains. 
Our data generation pipeline systematically controls task difficulty by varying the complexity and compositional structure of the editing intent.

Beyond proposing a more realistic and challenging task setting, we identify key limitations in existing evaluation paradigms, including LLM-as-a-judge frameworks and embedding-based similarity metrics such as CLIP~\cite{clip_paper}. 
LLM-judge approaches often produce scores within a narrow range, exhibit high variance under resampling, and provide limited interpretability regarding the source of errors~\cite{feng2025we, schroeder2024can, fu2025reliable}. 
Embedding-based methods, in contrast, tend to rely on coarse similarity signals and shallow heuristics, failing to capture semantically meaningful discrepancies between the generated and target visualizations~\cite{bai2025vision, li2025exploring}. 

To address these shortcomings, we propose a composite evaluation metric. 
Our framework first verifies that the generated code is executable and produces a valid visualization. 
It then applies structured assertions to assess adherence to user intent and compliance with code quality best practices. 
Finally, we introduce a grounded visual similarity component that leverages an LLM to explicitly enumerate visual differences between the generated and target charts. 
This scoring procedure yields fine-grained, interpretable feedback on model behavior while improving reliability through concrete constraints and explicit visual difference analysis.

\textbf{Our key contributions are:}
1. We formalize \emph{incremental chart code editing} as a benchmark task in which a model updates an existing chart specification given an initial code--plot pair and either a target visualization or a natural language description of the desired modification. \\
2. We introduce \textbf{ChartEditBench}, a difficulty-controlled synthetic dataset comprising 5{,}000 instances spanning 12 distinct modification types, along with a 430-example human-verified subset to support robust evaluation. \\
3. We propose visually grounded evaluation metrics that jointly measure rendering validity, visual similarity, and semantic correctness of code edits, yielding more reliable and interpretable assessments than generic LLM-as-a-judge approaches for analyzing chart editing behavior.

\section{Related Work}
\label{Chapter 2}

\subsection{Chart Editing and Intent-Aligned Visualization}

Most work on charts studies how models read information from a static image, rather than how they edit an existing visualization to better match user intent. Datasets such as ChartQA~\cite{masry2022chartqa}, PlotQA~\cite{methani2020plotqa}, and FigureQA~\cite{kahou2017figureqa} evaluate value extraction and reasoning over fixed charts, and models like DePlot~\cite{liu2022deplot} and MatCha~\cite{liu2022matcha} follow a plot-to-table pipeline that converts the chart into a textual table for downstream question answering. In real workflows, however, analysts and developers interact with charts by iteratively refining a shared specification, adjusting encodings, layouts, and statistical overlays while preserving useful structure and context, so the central question is whether a model can update the chart in a way that is consistent with evolving user intent rather than whether it can answer a single factual query. We therefore focus on \emph{chart editing} as a core capability for multimodal assistants and formalize it as editing an initial code–image pair into a new code–image pair that reflects the requested change while maintaining previously established decisions. To our knowledge, ChartEditBench is the first benchmark that explicitly targets this incremental chart editing setting, providing multi-step modification chains where success is defined by faithful state transitions across versions instead of one-shot chart understanding or generation~\cite{masry2022chartqa,methani2020plotqa,kahou2017figureqa,liu2022deplot,liu2022matcha}.

\subsection{From Single-Turn Generation to Agentic Chart Editors}

Recent chart-to-code and text-to-chart benchmarks focus on single-turn generation, where the model synthesizes plotting code from scratch or reconstructs code from a static target chart. MatPlotBench~\cite{chen2024matplotbench} evaluates natural language to visualization code, and Chart2Code~\cite{han2025chart2code} uses preference optimization to better match target visual styles, but both settings avoid the incremental scenario in which the model must modify an existing program while preserving its overall structure. In parallel, visual programming and multimodal agent work such as VisProg~\cite{gupta2023visual} and ViperGPT~\cite{suris2023vipergpt} shows that treating code as an intermediate representation enables more complex tool use and decomposition, and software engineering benchmarks like SWE-bench~\cite{jimenez2024swebench} demonstrate that multi-step, repository-level bug fixing reveals capabilities that are invisible in isolated code generation. ChartEditBench is designed in the same spirit for visualization: instead of repeatedly generating full scripts, models receive an initial chart specification and must apply a sequence of grounded edits, conditioned on either a natural language delta or a target chart, while carrying forward their own previous predictions. This multi-turn and stateful design makes ChartEditBench a role-equivalent benchmark for multimodal chart-editing agents, analogous to how SWE-bench serves as a realistic testbed for code agents in software engineering.

\subsection{Evaluation for Chart Editing}

Evaluation design is crucial for chart editing because we need scores that are reliable and interpretable at the level of individual edits. LLM-as-a-judge methods often show narrow and skewed score distributions, sensitivity to prompt and sampling choices, and limited transparency about which aspects of the output caused a low or high score, which makes it difficult to diagnose specific failure modes in a multi-turn editing process~\cite{lee2024improvingllmasjudge}. Embedding-based or CLIP-style metrics instead compare generated and reference images in a representation space~\cite{radford2021learning,huang2022evaluatingclipscore}, but they rely on coarse similarity signals and shallow heuristics that do not leverage the full reasoning capabilities of modern multimodal LLMs and are known to miss semantically important but visually subtle changes such as incorrect aggregations or misaligned axes. In ChartEditBench, we address these issues by combining strict code execution checks, structured assertions for instruction following and code quality, and a grounded visual similarity procedure in which an evaluation model must enumerate concrete visual differences under a fixed rubric . This composite, constraint-based framework provides finer-grained and more stable feedback than generic LLM judges or pure embedding metrics and aligns evaluation more closely with user-intent consistency in incremental chart editing~\cite{lee2024improvingllmasjudge,radford2021learning,huang2022evaluatingclipscore}.

\section{ChartEditBench: A Multi-Turn Chart Modification Benchmark}
\label{sec:ChartEditBench}

We introduce \textbf{ChartEditBench}, a synthetic benchmark designed to evaluate the incremental chart code editing capabilities of large multimodal models (LMMs). Unlike existing benchmarks that focus on single-turn chart reproduction or question answering, ChartEditBench emphasizes \textit{conversational chart modification}---a paradigm where models must iteratively refine visualization code across multiple turns while maintaining code-visual alignment.

\subsection{Task Definition}
\label{subsec:task_definition}

ChartEditBench defines two complementary tasks that reflect real-world chart editing workflows. In \textbf{Task 1: Code-to-Code Chart Modification}, given the original code $c_1$, the original chart image $p_1$, and a target modified chart image $p_2$, the model must predict the modified code $c_2$ that produces $p_2$. Formally, $c_2 = f(c_1, p_1, p_2)$. This task evaluates the model's ability to reason about visual differences between charts and translate them into precise code modifications. In \textbf{Task 2: Natural Language Delta Modification}, given the original code $c_1$, the original chart image $p_1$, and a natural language instruction $\delta_{nl}$ describing the desired modification, the model must generate the modified code $c_2$, where $c_2 = f(c_1, p_1, \delta_{nl})$. This task assesses instruction-following capabilities in the context of incremental code editing. Both tasks are evaluated across multi-turn conversations, where each subsequent turn builds upon the model's previous prediction, creating a chain of dependent modifications that tests long-horizon reasoning and error accumulation.

\subsection{Dataset Construction}
\label{subsec:dataset_construction}

\subsubsection{Synthetic Data Generation Pipeline}

We employ a fully automated pipeline leveraging large language models to generate diverse, executable chart code and corresponding modification sequences. The generation process consists of three stages. In \textbf{Stage 1: Initial Chart Generation}, we generate an initial chart by prompting an LLM with structured specifications including chart type (selected from 37 distinct matplotlib chart types spanning basic plots, statistical visualizations, specialized charts, 3D plots, and polar coordinates), data specifications (number of data points, series count, and value ranges with fixed random seeds for reproducibility), and visual parameters (color schemes, themes, figure dimensions, font sizes, and grid configurations). The generated code is executed in a sandboxed Python environment, and only instances producing valid, renderable charts proceed to subsequent stages.

In \textbf{Stage 2: Multi-Turn Modification Generation}, each instance undergoes $N$ sequential modification turns (default $N=5$), with difficulty progressively increasing. Turn 1 focuses on simple styling changes (colors, titles, labels), Turns 2--3 introduce structural modifications (markers, line styles, axis scaling), and Turns 4--5 require complex transformations (chart type conversion, subplot arrangement, dual axes). For each turn, the LLM receives the previous code and generates a modification instruction along with the corresponding modified code, both of which are executed and validated before proceeding. In \textbf{Stage 3: VQA Generation}, we generate 5 visual question-answering pairs for each rendered chart using LLM prompting, covering chart type identification, visual element properties, data interpretation, and structural analysis. This enables future evaluation of chart understanding capabilities.

\subsubsection{Uniform Distribution Management}

To ensure balanced coverage across chart types and modification categories, we implement a \textbf{UniformDistributionManager}. Chart types are organized in a queue and selected cyclically using a round-robin strategy, ensuring each of the 37 chart types receives approximately equal representation across the dataset. We categorize modifications into two families: \textit{style modifications} (17 types) encompassing visual changes such as color schemes, typography, markers, line styles, themes, and transparency adjustments, and \textit{data modifications} (18 types) involving structural changes including data filtering, axis scaling, aggregation, trend lines, chart type conversion, and subplot arrangements. The system alternates between style and data modifications across turns, with each modification type drawn from its respective queue using round-robin selection, ensuring balanced coverage of both categories.

\subsubsection{Code Execution and Validation}

All generated code undergoes rigorous validation through a four-step process: syntax validation via Abstract Syntax Tree (AST) parsing to ensure syntactic correctness, execution validation through code execution in isolated subprocesses with 60-second timeouts, render validation to verify that \texttt{plt.savefig()} produces a valid image file, and structure validation to confirm the presence of required elements (matplotlib imports, savefig call, close statement). Instances failing any validation step are discarded, and generation is retried with exponential backoff (up to 3 attempts).

\subsection{Dataset Verification}
\label{subsec:verification}

\subsubsection{Automated Verification with Chart-R1}

We employ Chart-R1~\cite{chart-r1}, a vision-language model fine-tuned for chart understanding, to perform automated quality assessment. For each modification turn $t > 0$, we conduct two types of verification. In \textit{code-to-code verification}, given the original code $c_{t-1}$, modified code $c_t$, and modification instruction, Chart-R1 assesses whether the code changes correctly implement the instruction, returning a binary judgment. In \textit{chart-to-chart verification}, given the original chart image $p_{t-1}$, modified chart image $p_t$, and modification instruction, Chart-R1 evaluates whether the visual changes align with the instruction.

\subsubsection{Two-Type Assertion System}

To enable fine-grained programmatic evaluation, we generate two categories of assertions for each modification. \textit{Instruction-following assertions} (3--4 per turn) verify that the specific modification instruction was implemented correctly. For example, for the instruction ``Change markers to '*' with size 8'', we generate assertions checking for \texttt{marker='*'} and \texttt{markersize=8} in the code. \textit{Requirement assertions} (4--8 per turn) verify code quality and structural correctness independent of the specific instruction, such as the presence of matplotlib imports, \texttt{plt.savefig()}, and \texttt{plt.close()} statements. This dual-assertion approach enables both instruction-specific and general code quality evaluation.

\subsubsection{Instruction Classification}

Each modification instruction is classified into one of two verification types based on its evaluability. \textit{Programmatic instructions} have objectively verifiable outcomes (e.g., ``change color to red'', ``set axis limit to 100'') and are evaluated using assertion pass rates. \textit{LLM-judged instructions} require semantic interpretation (e.g., ``make the chart more professional'', ``improve readability'') and are evaluated using LLM-based scoring. This classification enables hybrid evaluation that leverages the strengths of both programmatic and neural assessment.

\subsubsection{Human Verification}

To validate automated assessment quality, we uniformly sample 400 instances across chart types and conduct human verification with two expert annotators. The sampling procedure ensures proportional representation, where $n_{\text{type}} = \lfloor \frac{|\mathcal{D}_{\text{type}}|}{|\mathcal{D}_{\text{total}}|} \times 400 \rfloor$. Annotators independently evaluate each modification turn for code correctness relative to the instruction, visual fidelity of rendered charts, and preservation of unmodified elements. Inter-annotator agreement is measured using Cohen's $\kappa$, and disagreements are resolved through discussion.

\subsection{Evaluation Framework}
\label{subsec:evaluation}

\subsubsection{Conversational Benchmarking Protocol}

Unlike single-turn evaluation, ChartEditBench employs a \textbf{conversational protocol} where each turn uses the model's own previous prediction as input. At Turn 0$\rightarrow$1, the model receives ground-truth $c_1$, $p_1$, and target $p_2$ (Task 1) or instruction $\delta_{nl}$ (Task 2), and predicts $\hat{c}_2$. At Turn 1$\rightarrow$2, the model receives its own $\hat{c}_2$, the rendered $\hat{p}_2$, and the next target or instruction, then predicts $\hat{c}_3$. Subsequent turns continue building on model predictions. This protocol reveals error accumulation effects and tests robustness to imperfect intermediate states. If a model's predicted code fails to render at turn $t$, the system automatically falls back to the last successfully rendered code and chart from turn $t-k$ (where $k \geq 1$), tracked by the \texttt{using\_fallback} flag.

\subsubsection{Evaluation Metrics}

We report five complementary metrics. \textit{Execution Rate} measures the percentage of predicted code samples that execute without errors and produce valid chart images: $\text{Exec. Rate} = \frac{\#\text{ successfully rendered}}{\#\text{ total predictions}}$. \textit{Instruction Following Score} quantifies adherence to modification instructions. For programmatic instructions, this is the pass rate of instruction-following assertions: $\text{IF}_{\text{prog}} = \frac{\#\text{ assertions passed}}{\#\text{ total assertions}}$. For LLM-judged instructions, this is the LLM evaluation score (1--5 scale, normalized to 0--1). \textit{Code Quality Score} measures the pass rate of requirement assertions, assessing structural correctness independent of modification content: $\text{CQ} = \frac{\#\text{ requirement assertions passed}}{\#\text{ total requirement assertions}}$. \textit{Visual Similarity Score} provides LLM-based comparison between predicted and ground-truth charts, identifying major differences (1.0 point deduction each) and minor differences (0.5 point deduction each): $\text{VS} = \max(0, 5.0 - 1.0 \times n_{\text{major}} - 0.5 \times n_{\text{minor}})$. Finally, \textit{Overall Score} averages instruction following and visual similarity, normalized to a common scale.

\subsection{Dataset Statistics}
\label{subsec:statistics}

The final ChartEditBench dataset comprises 4,142 total instances, with 6 turns per instance (1 initial + 5 modifications) yielding 20,710 total modification samples. The dataset covers 37 distinct chart types across 6 categories and 35 distinct modification types (17 style, 18 data), with a human-verified subset of 400 instances (2,400 turns). The distribution ensures approximately equal representation across chart types and balanced coverage of style versus data modifications, maintaining an approximately 50/50 split across the dataset. 
\section{Experiments}
\label{sec:experiments}

We conduct comprehensive experiments to evaluate state-of-the-art large multimodal models on ChartEditBench across both Task 1 (Code-to-Code) and Task 2 (NL Delta) settings.

\subsection{Experimental Setup}
\label{subsec:experimental_setup}

\subsubsection{Models}

We evaluate a diverse set of proprietary and open-source LMMs spanning different model scales and architectures. For proprietary models, we test GPT-5-mini~\cite{openai2025}, OpenAI's compact multimodal model with strong vision-language capabilities, and Claude Haiku 4.5~\cite{anthropic2025}, Anthropic's efficient model variant accessed via the CMU AI Gateway\footnote{\url{https://ai-gateway.andrew.cmu.edu/}}. For open-source models, we evaluate Qwen3-VL~\cite{qwenvl2024} (2B, 8B, 30B-A3B), Alibaba's vision-language models with varying capacities where the 30B variant uses a Mixture-of-Experts architecture with 3B active parameters, and InternVL3~\cite{internvl2025} (1B, 8B), open-source VLMs from Shanghai AI Laboratory with strong chart understanding capabilities.

\subsubsection{Inference Configuration}

We access GPT-5-mini and Claude Haiku 4.5 with maximum completion tokens set to 8,192 for standard turns and 32,768 for complex modifications, and temperature fixed at 0.1 for reproducibility. All open-source models are served using vLLM~\cite{kwon2023vllm} on NVIDIA A100 80GB GPUs (1 GPU for models $\leq$8B, 2 GPUs with tensor parallelism for 30B models), configured with 85--90\% memory utilization, maximum model length of 16,384 tokens, and eager execution mode with chunked prefill disabled. Images are processing without any transformation using each model's maximum supported input pixel configuration to preserve chart details.

\subsubsection{Evaluation Models}

For LLM-based evaluation of instruction following and visual similarity, we employ two models. Chart-R1~\cite{chart-r1}, a Qwen2.5-VL-based model fine-tuned specifically for chart understanding and evaluation, serves as our primary evaluation model for both code-to-code and chart-to-chart assessment. GPT-5-mini is used as an alternative evaluator for visual similarity scoring, following the LMM-score methodology from prior work~\cite{tang2025charts}.

\subsection{Benchmarking Protocol}
\label{subsec:benchmarking_protocol}

\subsubsection{Conversational Evaluation}

We implement the conversational protocol described in Section~\ref{subsec:evaluation}, where each model's predictions become the input for subsequent turns. This creates a chain of dependent modifications: $\hat{c}_{t+1} = f_\theta(\hat{c}_t, \hat{p}_t, I_{t+1})$, where $\hat{c}_t$ and $\hat{p}_t$ are the model's predicted code and rendered chart from the previous turn, and $I_{t+1}$ is either the target chart image (Task 1) or natural language instruction (Task 2). For each instance, we maintain three state variables: \texttt{last\_successful\_code} (the most recent code that rendered successfully), \texttt{last\_successful\_plot} (the corresponding rendered chart), and \texttt{using\_fallback} (a boolean flag indicating whether fallback was triggered). When a model's predicted code fails to render, subsequent turns automatically use the last successful state, preventing complete failure cascades while accurately tracking where models break.

\subsubsection{Output Processing}

Model outputs are parsed to extract Python code using a hierarchical extraction strategy. We first attempt to parse JSON responses with \texttt{code}, \texttt{explanation}, and \texttt{modifications} fields, then extract code from \texttt{```python ... ```} markdown blocks, followed by generic \texttt{``` ... ```} blocks, and finally treat the entire response as code if no delimiters are found. Extracted code is post-processed to ensure the presence of \texttt{plt.savefig('plot.png', dpi=300, bbox\_inches='tight')}, \texttt{plt.close()} for memory management, and correct output paths for rendered charts.

\subsection{Evaluation Procedure}
\label{subsec:evaluation_procedure}

\subsubsection{Assertion-Based Evaluation}

For each modification turn, we execute both assertion categories. For instruction-following assertions, we execute the 3--4 instruction-specific assertions against the predicted code string, where the pass rate directly becomes the instruction-following score for programmatic instructions: $\text{IF}_{\text{prog}} = \frac{1}{|A_{if}|} \sum_{a \in A_{if}} \mathbf{1}[\text{exec}(a, \hat{c}) = \texttt{True}]$. For requirement assertions, we execute the 4--8 code quality assertions, providing the code quality score: $\text{CQ} = \frac{1}{|A_{req}|} \sum_{a \in A_{req}} \mathbf{1}[\text{exec}(a, \hat{c}) = \texttt{True}]$.

\subsubsection{LLM-Based Evaluation}

For LLM-judged instructions and visual similarity, we employ Chart-R1 with structured prompts. For instruction following, Chart-R1 receives the original code, predicted code, ground-truth code, modification instruction, and three chart images (original, ground-truth modified, predicted modified), returning a 1--5 score assessing semantic correctness. For visual similarity, Chart-R1 compares the ground-truth and predicted charts, identifying major differences (wrong chart type, missing data series, incorrect scale types, missing legends) and minor differences (color variations, font size discrepancies, spacing differences, gridline variations). The visual similarity score is computed as $\text{VS} = \max(0, 5 - n_{\text{major}} - 0.5 \times n_{\text{minor}})$.

\subsubsection{Hybrid Scoring}

The final instruction-following score is determined by the instruction classification: $\text{IF} = \begin{cases} \text{IF}_{\text{prog}} & \text{if classification} = \texttt{programmatic} \\ \text{IF}_{\text{llm}} / 5 & \text{if classification} = \texttt{llm} \end{cases}$. This ensures that objective instructions receive objective evaluation while subjective instructions receive appropriate semantic assessment.

\subsection{Experimental Configurations}
\label{subsec:configurations}

We evaluate all models on several configurations to provide comprehensive analysis. For full dataset evaluation, all 4,142 instances (24,852 total turns including initial charts) are processed for each model-task combination, providing comprehensive coverage across chart types and modification difficulties. The 400 uniformly sampled instances (2,400 turns) with human annotations serve as a high-confidence evaluation subset for detailed analysis. Results are disaggregated by turn number to analyze error accumulation across the conversation, difficulty progression effects, and fallback frequency at different turns. Results are further disaggregated by chart type (37 categories), modification category (style vs. data), modification difficulty (easy, medium, hard, expert), and verification type (programmatic vs. LLM-judged).

\subsection{Computational Resources}
\label{subsec:resources}

All experiments are conducted on a compute cluster with NVIDIA A100 80GB GPUs, 16 CPU cores per node, 256GB RAM per node, and 120 hours time allocation per benchmark run. For each model-task pair, small models (1B--2B) require a single A100 and approximately 8--12 hours for the full dataset, medium models (7B--8B) require a single A100 and approximately 18--24 hours, while large models (30B+) require dual A100 with tensor parallelism and approximately 36--48 hours. Chart-R1 evaluation runs as a separate job after benchmarking, requiring an additional 12--24 hours per model depending on the number of successfully rendered predictions.

\subsection{Reproducibility}
\label{subsec:reproducibility}

To ensure reproducibility, we implement several measures. All chart code uses fixed random seeds (\texttt{np.random.seed(42)}) for data generation, ensuring identical charts across runs. The benchmarking system supports automatic resume from interruptions, scanning output directories for completed instances and continuing from the last successful checkpoint. Each prediction is logged with full model response and extracted code, execution stdout/stderr, rendering success/failure status, timing information, and fallback usage indicators. All rendered charts, predicted code files, and intermediate results are preserved in a structured directory hierarchy where each instance contains organized subdirectories for different turns, storing original, ground-truth, and predicted charts alongside their corresponding code files. This enables post-hoc analysis and human inspection of model behavior.

\section{Results}
\label{sec:results}

We evaluate six models on \textsc{ChartEditBench}, analyzing 400 evaluated samples across 2400 total turns. Our results reveal significant performance gaps between model scales and architectures, consistent degradation patterns across conversation turns, and systematic failure modes on specific modification types.

\subsection{Overall Performance}
\label{subsec:overall_performance}

Table~\ref{tab:main_results} presents the main results across all models. Claude Haiku 4.5 achieves the highest overall score (2.187), demonstrating strong performance across all metrics with particularly high visual similarity (1.937). GPT-5-mini leads in instruction following (1.623) but shows weaker visual similarity (1.521), suggesting it prioritizes explicit instruction compliance over maintaining visual fidelity. Among open-source models, Qwen3-VL-30B-A3B achieves competitive performance (2.139 overall) despite using only 3B active parameters through its MoE architecture, while the 8B variant shows strong visual similarity (1.964) but moderate instruction following (1.186).

\begin{table}[t]
\centering
\caption{Main results on \textsc{ChartEditBench}. IF = Instruction Following (0--2 scale), CQ = Code Quality (0--1 scale), VS = Visual Similarity (0--5 scale), Overall = weighted composite (0--5 scale). Best results in \textbf{bold}, second-best \underline{underlined}. All scores computed as weighted means on successfully evaluated samples.}
\label{tab:main_results}
\resizebox{\columnwidth}{!}{%
\begin{tabular}{lcccc}
\toprule
\textbf{Model} & \textbf{IF} $\uparrow$ & \textbf{CQ} $\uparrow$ & \textbf{VS} $\uparrow$ & \textbf{Overall} $\uparrow$ \\
\midrule
\multicolumn{5}{l}{\textit{Proprietary Models}} \\
GPT-5-mini & \textbf{1.623} & 0.763 & 1.521 & 2.094 \\
Claude Haiku 4.5 & \underline{1.397} & \underline{0.777} & 1.937 & \textbf{2.187} \\
\midrule
\multicolumn{5}{l}{\textit{Open-Source Models}} \\
Qwen3-VL-30B-A3B & 1.249 & \textbf{0.783} & \textbf{2.072} & \underline{2.139} \\
Qwen3-VL-8B & 1.186 & 0.767 & \underline{1.964} & 2.045 \\
Qwen3-VL-2B & 0.721 & 0.627 & 1.482 & 1.426 \\
InternVL3-1B & 0.480 & 0.691 & 1.791 & 1.345 \\
\bottomrule
\end{tabular}%
}
\end{table}

The smallest models exhibit substantial performance gaps. InternVL3-1B achieves only 0.480 instruction following, indicating fundamental limitations in understanding and executing chart modification instructions. Interestingly, InternVL3-1B maintains relatively high visual similarity (1.791), suggesting it can preserve chart structure even when failing to implement specific modifications---likely because failed modifications result in minimal changes rather than corrupted outputs.

\subsection{Performance Degradation Across Turns}
\label{subsec:turn_analysis}

\begin{figure}[t]
\centering
\includegraphics[width=\columnwidth]{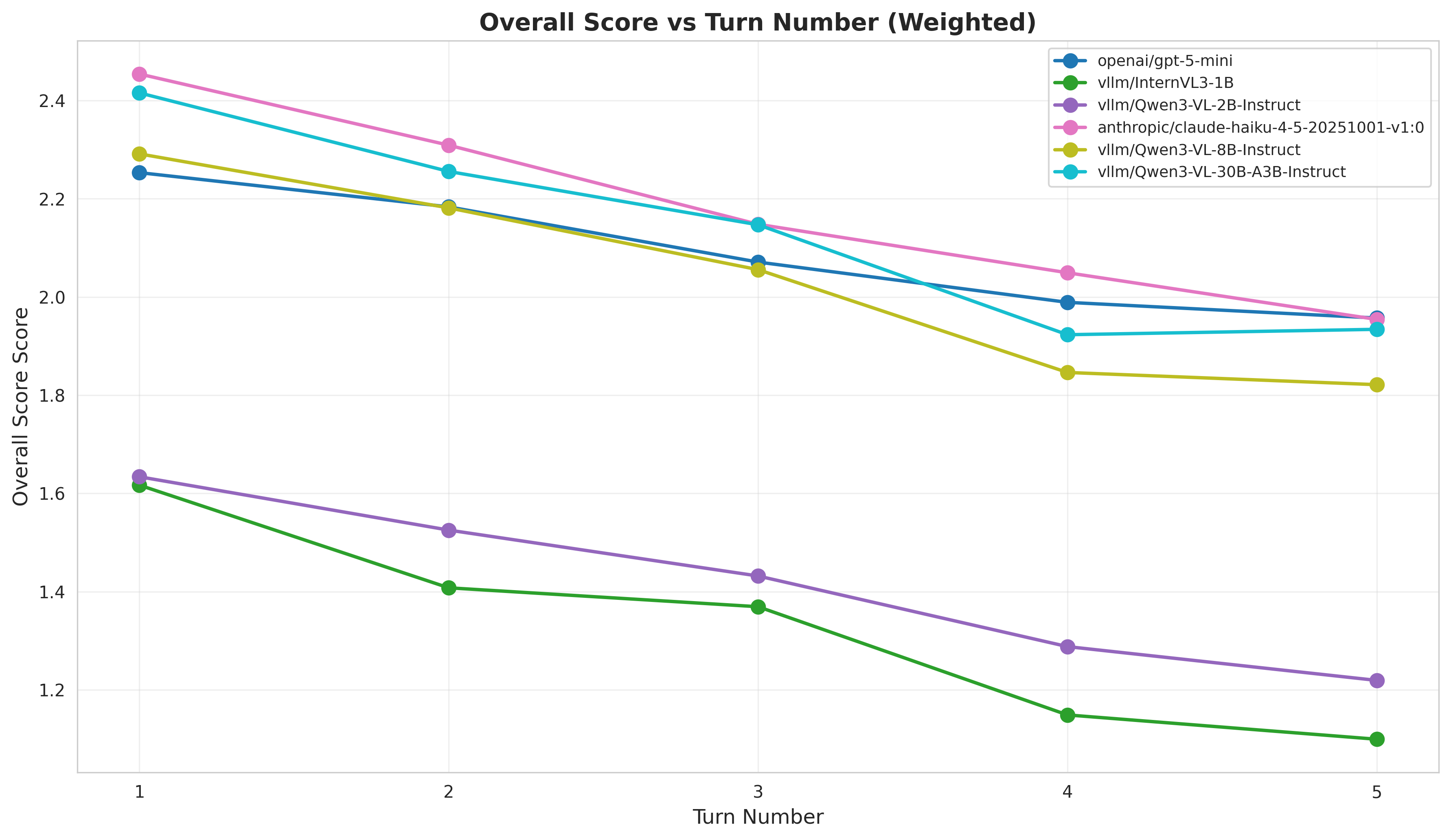}
\caption{Overall score degradation across conversation turns. All models exhibit declining performance, with smaller models (InternVL3-1B, Qwen3-VL-2B) showing steeper drops. The gap between high-performing and low-performing models widens progressively.}
\label{fig:turn_overall}
\end{figure}

A key finding of our evaluation is the consistent performance degradation across conversation turns, illustrated in Figure~\ref{fig:turn_overall}. This error accumulation effect is fundamental to understanding real-world multi-turn chart editing scenarios.

All models show monotonic performance decline from turn 1 to turn 5. Claude Haiku 4.5 degrades from 2.46 to 1.97 (20\% drop), while InternVL3-1B drops from 1.62 to 1.09 (33\% drop). This demonstrates that error accumulation disproportionately affects weaker models, wherein early mistakes compound more severely when the model cannot recover. The performance gap between the best and worst models increases from 0.84 at turn 1 to 0.88 at turn 5, indicating that multi-turn evaluation amplifies capability differences.

Figure~\ref{fig:turn_metrics} in the Appendix provides disaggregated analysis by metric. Visual similarity shows the steepest decline (approximately 35\% average drop), as accumulated code errors increasingly corrupt chart appearance. Code quality degrades more gradually (approximately 8\% drop), since structural code properties are more robust to modification errors. Instruction following remains relatively stable across turns, suggesting that models' ability to understand instructions is independent of accumulated state corruption.

\subsection{Difficulty-Stratified Analysis}
\label{subsec:difficulty_analysis}

\begin{figure}[t]
\centering
\includegraphics[width=\columnwidth]{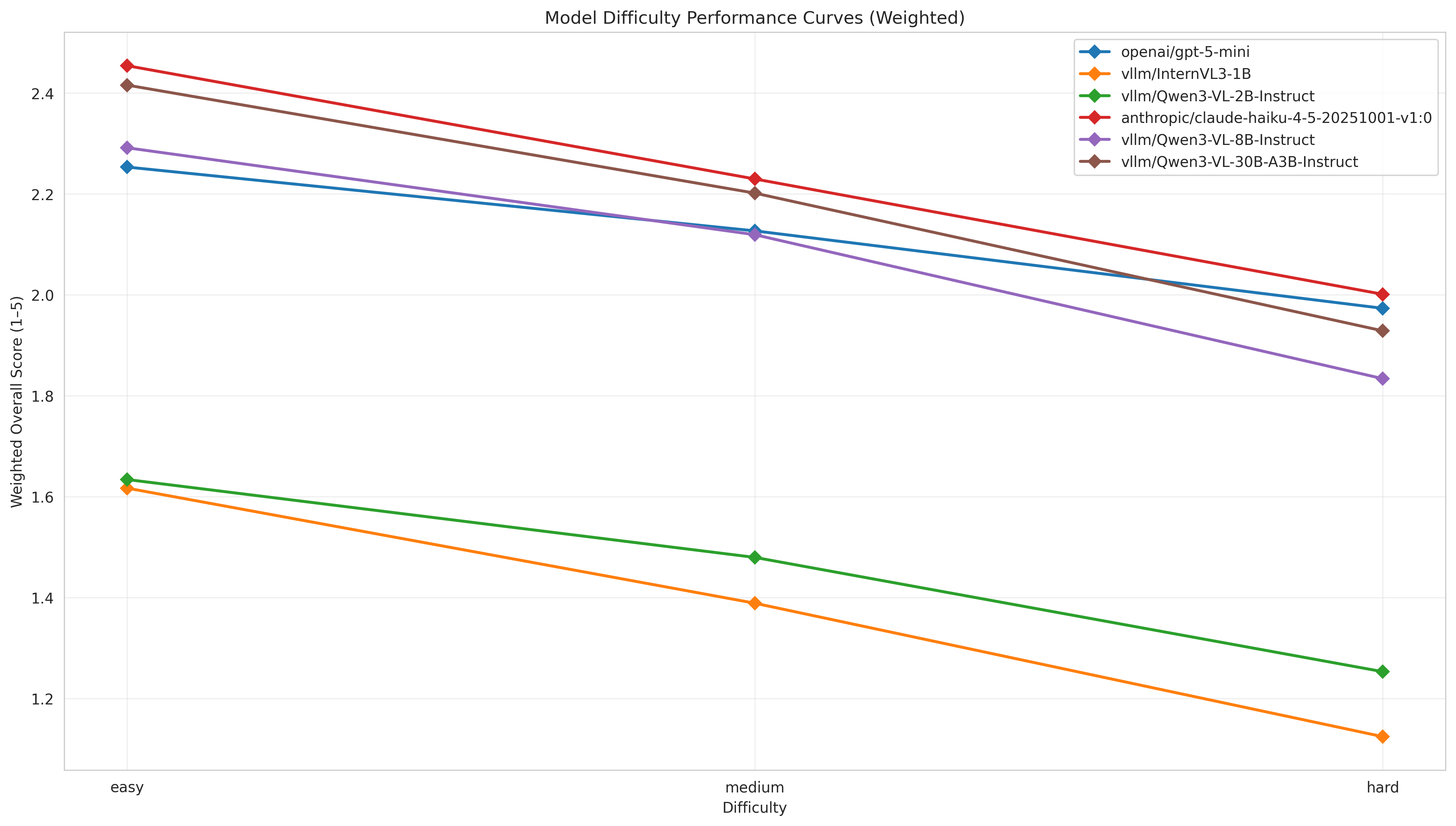}
\caption{Model performance curves across difficulty levels (weighted by rendering success). Proprietary models and larger open-source models maintain performance better on hard modifications, while smaller models show steeper degradation.}
\label{fig:difficulty_curves}
\end{figure}

Figure~\ref{fig:difficulty_curves} shows performance curves across difficulty levels. All models exhibit expected performance degradation from easy to hard modifications, but the rate of decline varies significantly by model capability.

Claude Haiku 4.5 demonstrates the most graceful degradation, dropping from 2.45 (easy) to 2.00 (hard)---an 18\% relative decline. In contrast, InternVL3-1B degrades from 1.62 to 1.11 (31\% decline), indicating that smaller models disproportionately struggle with complex modifications. Notably, the performance ordering among top models remains consistent across all difficulty levels, suggesting that \textsc{ChartEditBench} difficulty calibration effectively stratifies model capabilities.

An interesting finding emerges from the instruction following metric (see Appendix Figure~\ref{fig:difficulty_if}): GPT-5-mini and Claude Haiku 4.5 show \textit{increasing} instruction following scores at higher difficulty levels. This counterintuitive result occurs because harder modifications tend to involve more LLM-judged (semantic) instructions, where these models excel, rather than programmatic assertions where pass/fail is binary.

\subsection{Modification Type Analysis}
\label{subsec:modification_analysis}

\begin{figure*}[!tbh]
\centering
\includegraphics[width=2\columnwidth]{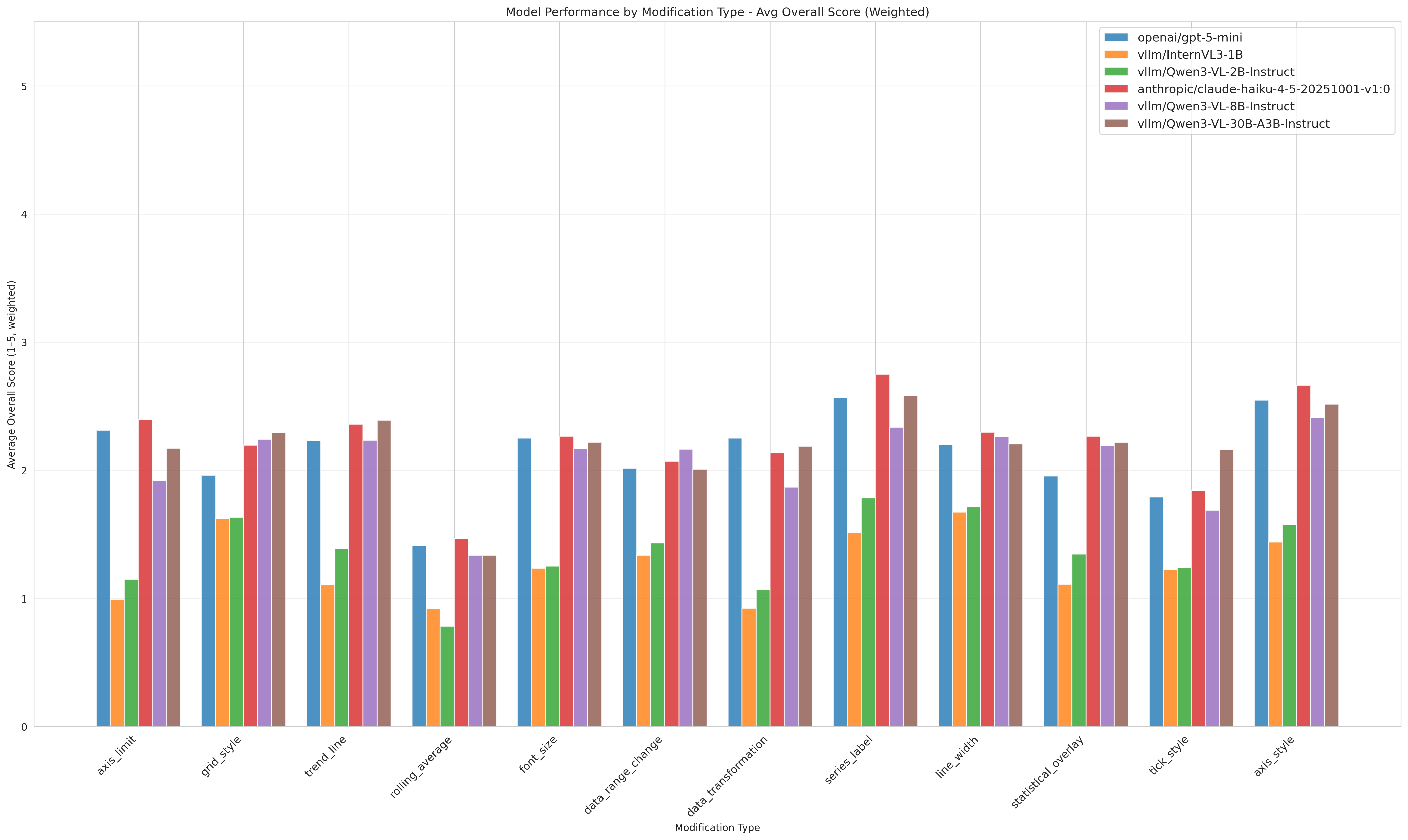}
\caption{Model performance by modification type. Style-focused modifications (axis\_style, series\_label) yield higher scores, while data-centric operations (rolling\_average, data\_transformation) prove more challenging across all models.}
\label{fig:modification_scores}
\end{figure*}

Performance varies substantially across modification types, revealing systematic strengths and weaknesses. Figure~\ref{fig:modification_scores} presents weighted average scores for the most frequent modification categories.

Style modifications such as \texttt{axis\_style}, \texttt{series\_label}, and \texttt{axis\_limit} yield the highest scores across all models (2.2--2.8 for top performers), as these require straightforward parameter changes without complex data manipulation. Conversely, data-centric modifications like \texttt{rolling\_average} and \texttt{data\_transformation} prove challenging even for the best models (1.3--1.5 average), requiring correct implementation of mathematical operations on data structures.

The \texttt{rolling\_average} modification emerges as particularly difficult, with all models achieving below 1.5 average score. Error analysis reveals that models frequently make off-by-one errors in window calculations, incorrectly handle edge cases, or apply averaging to wrong data series. This highlights a fundamental limitation in current LMMs' ability to reason about sequential data operations.

\subsection{Programmatic vs.\ LLM-Judged Instructions}
\label{subsec:eval_type_analysis}

\begin{figure*}[!tbh]
\centering
\includegraphics[width=2\columnwidth]{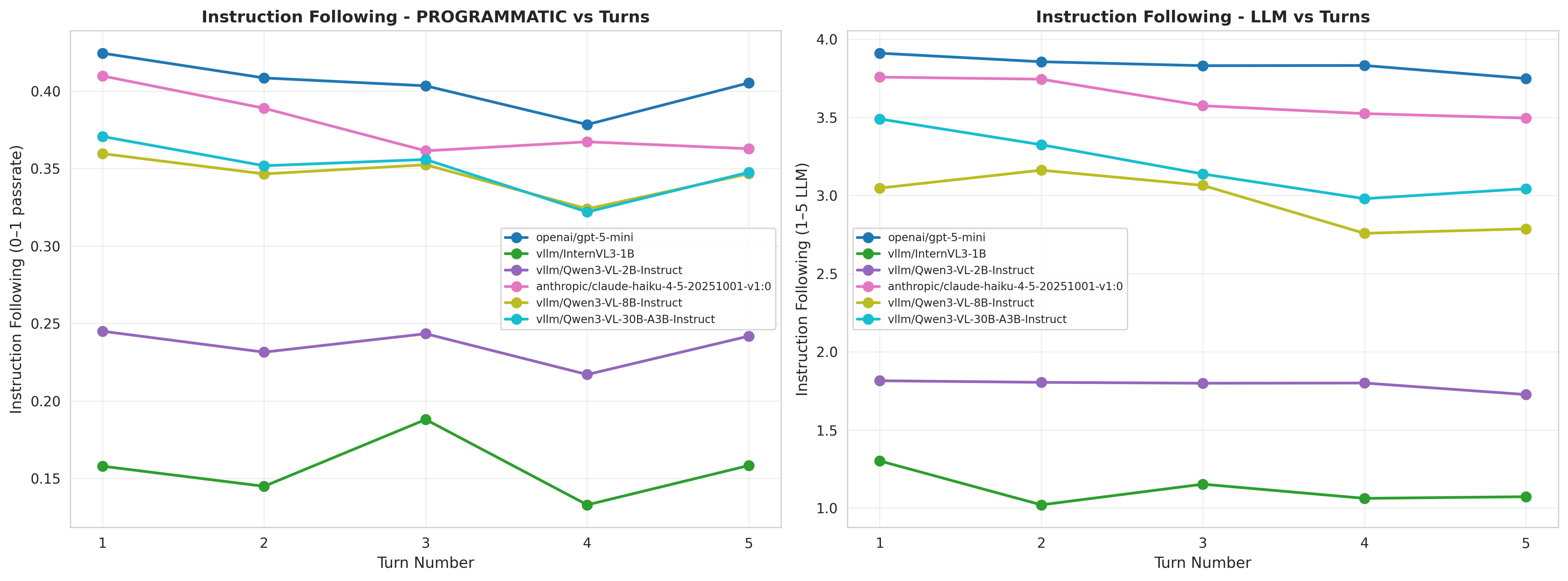}
\caption{Instruction following by evaluation type across turns. Left: Programmatic (assertion-based, 0--1 scale). Right: LLM-judged (semantic, 1--5 scale). Top models maintain advantage in both categories, with LLM-judged scores showing more stability across turns.}
\label{fig:instruction_classes}
\end{figure*}

Our hybrid evaluation reveals distinct performance patterns between programmatic and LLM-judged instructions (Figure~\ref{fig:instruction_classes}). GPT-5-mini achieves the highest programmatic pass rate (approximately 0.42), demonstrating strong ability to implement precisely specified modifications. Claude Haiku 4.5 follows closely (0.41), while smaller models struggle significantly with programmatic assertions (InternVL3-1B: 0.16).

For LLM-judged instructions, the gap between models narrows somewhat, with scores ranging from 3.0--4.0 for top models versus 1.0--2.0 for smaller models. This suggests that semantic understanding of modification intent is more evenly distributed than precise code implementation ability. The stability of LLM-judged scores across turns (compared to programmatic scores) indicates that models' understanding of \textit{what} to do remains intact even as their ability to \textit{correctly implement} modifications degrades.

\subsection{Rendering Success Analysis}
\label{subsec:rendering_analysis}

Beyond quality metrics, we analyze rendering success rates---the percentage of model outputs that produce valid, executable matplotlib code. Proprietary models achieve 85--90\% rendering success across modification types, while open-source models range from 46\% (Qwen3-VL-2B on rolling\_average) to 90\% (Qwen3-VL-30B on trend\_line). 

The modification type significantly impacts rendering success (see Appendix Figure~\ref{fig:modification_success}). \texttt{Trend\_line} modifications achieve nearly 100\% success for GPT-5-mini and Claude Haiku 4.5, as these involve straightforward additions to existing plots. In contrast, \texttt{rolling\_average} causes frequent rendering failures due to index misalignment and data shape mismatches. This correlation between modification complexity and rendering success suggests that code validity is a prerequisite challenge before quality optimization.

\subsection{Key Findings and Implications}
\label{subsec:key_findings}

Our comprehensive evaluation yields several actionable findings:

\paragraph{Error Accumulation is Fundamental.} Multi-turn evaluation reveals capabilities invisible to single-turn benchmarks. The 20--33\% performance drop from turn 1 to turn 5 demonstrates that real-world iterative editing scenarios pose substantially greater challenges than isolated modifications.

\paragraph{Model Scale Matters, But Efficiency Helps.} Qwen3-VL-30B-A3B's competitive performance (2.139 overall) with only 3B active parameters suggests that MoE architectures offer promising efficiency-capability trade-offs for chart understanding tasks.

\paragraph{Data Operations Remain Challenging.} Even the best models struggle with modifications requiring data manipulation (rolling averages, transformations), highlighting a gap between visual understanding and numerical reasoning that future models must address.

\paragraph{Hybrid Evaluation Captures Complementary Dimensions.} The divergence between programmatic and LLM-judged performance (Figure~\ref{fig:instruction_classes}) validates our hybrid approach---models that excel at precise implementation may underperform on semantic modifications, and vice versa.

\section{Conclusion} 
\label{sec:conclusion}

We introduced ChartEditBench, a benchmark for evaluating incremental chart code editing in multimodal large language models. The benchmark comprises 5,000 synthetic instances spanning 35 modification types, with a 430-sample human-verified subset ensuring unambiguous ground truth. Our visually grounded evaluation metrics---combining rendering success, visual similarity, and code-level correctness---provide reliable assessment while avoiding calibration issues inherent in unconstrained LLM judging.

Experiments across six models reveal three key findings: (1) strong capacity-dependent performance gaps, with Claude Haiku 4.5 achieving 2.187 overall score versus 1.345 for InternVL3-1B; (2) systematic degradation of 20--33\% across multi-turn conversations, amplifying capability differences between models; and (3) persistent challenges in data-centric modifications such as rolling averages, where even top models struggle. These results demonstrate that current MLLMs, while proficient at single-turn generation, face substantial limitations in iterative, visually grounded code editing workflows.

ChartEditBench establishes a challenging testbed for developing conversational chart-editing agents and probing the boundaries of multimodal reasoning over code, images, and instructions.

\bibliography{custom}
\appendix

\section{Additional Experimental Details}
\label{app:experimental_details}

\subsection{Dataset Statistics}
\label{app:dataset_stats}

\begin{figure}[h]
\centering
\includegraphics[width=0.8\columnwidth]{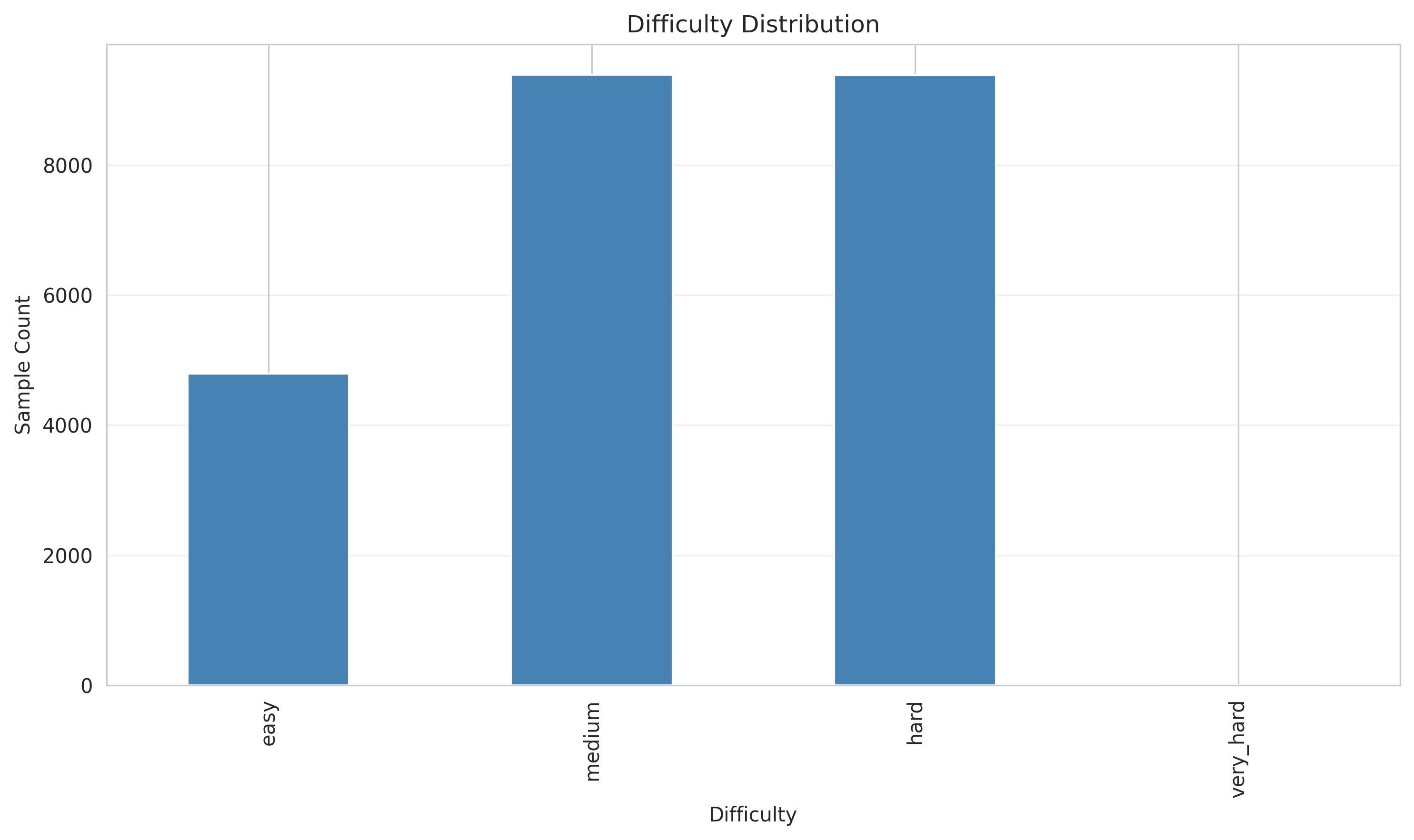}
\caption{Distribution of samples across difficulty levels in \textsc{ChartEditBench}. The dataset emphasizes medium and hard modifications to maximize evaluation discriminability.}
\label{fig:difficulty_dist}
\end{figure}

Figure~\ref{fig:difficulty_dist} presents the distribution of samples across difficulty levels in \textsc{ChartEditBench}. The dataset is intentionally weighted toward medium and hard modifications to provide discriminative evaluation, with approximately 4,800 easy samples, 9,400 medium samples, and 9,400 hard samples. Very hard modifications are rare by design, reserved for expert-level multi-step operations.

\begin{figure}[h]
\centering
\includegraphics[width=0.9\columnwidth]{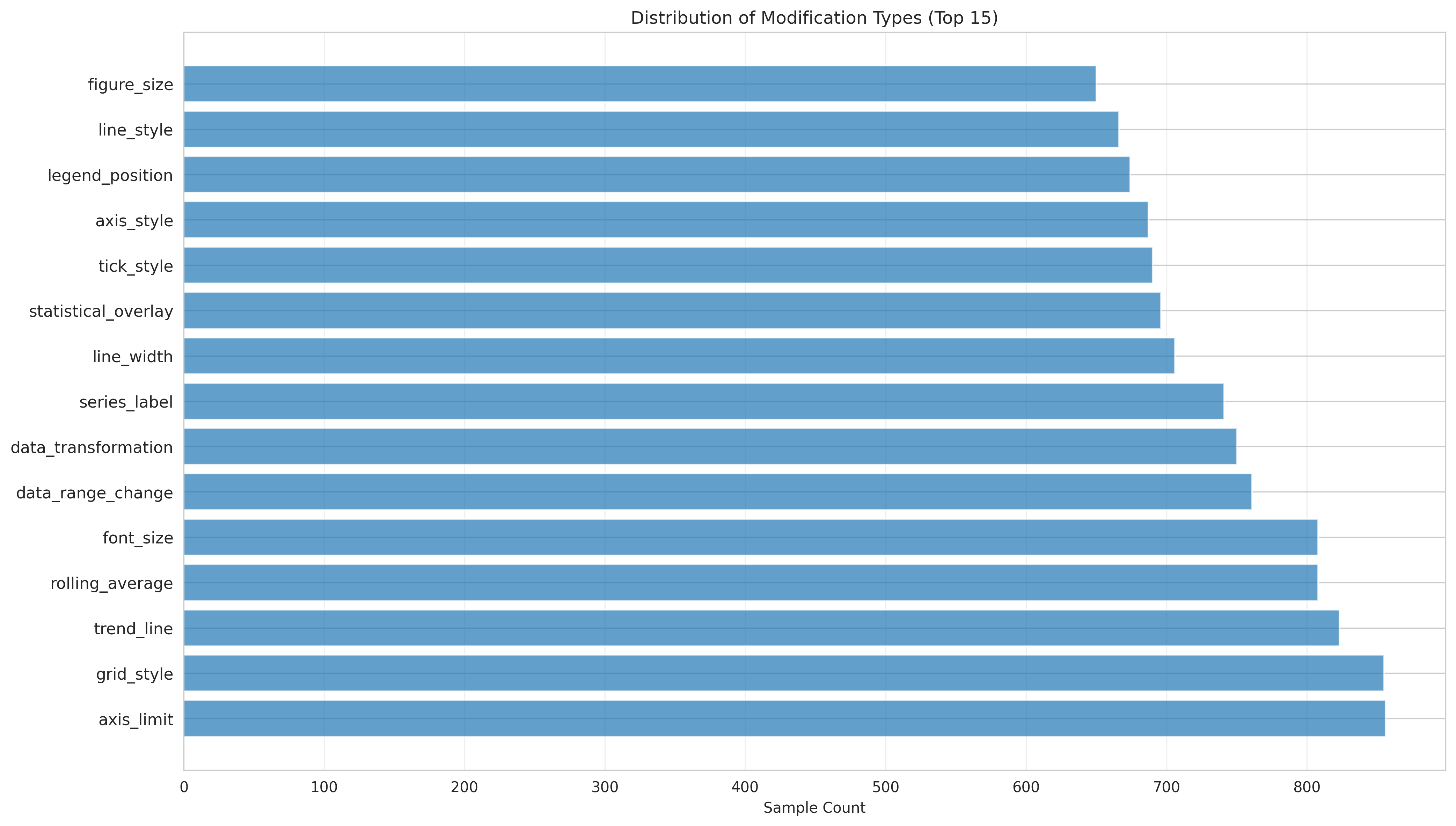}
\caption{Distribution of modification types (top 15 categories). The benchmark balances style modifications with data-centric operations.}
\label{fig:modification_dist}
\end{figure}

Figure~\ref{fig:modification_dist} shows the distribution of the top 15 modification types. The benchmark covers a diverse range of operations, with axis-related modifications (\texttt{axis\_limit}, \texttt{axis\_style}) and styling operations (\texttt{grid\_style}, \texttt{trend\_line}) being most frequent. Data manipulation operations (\texttt{data\_transformation}, \texttt{data\_range\_change}, \texttt{rolling\_average}) are well-represented to ensure comprehensive evaluation of numerical reasoning capabilities.

\section{Extended Results}
\label{app:extended_results}

\subsection{Detailed Turn-Based Analysis}
\label{app:turn_analysis}

\begin{figure*}[t]
\centering
\begin{subfigure}[t]{0.48\textwidth}
\centering
\includegraphics[width=\textwidth]{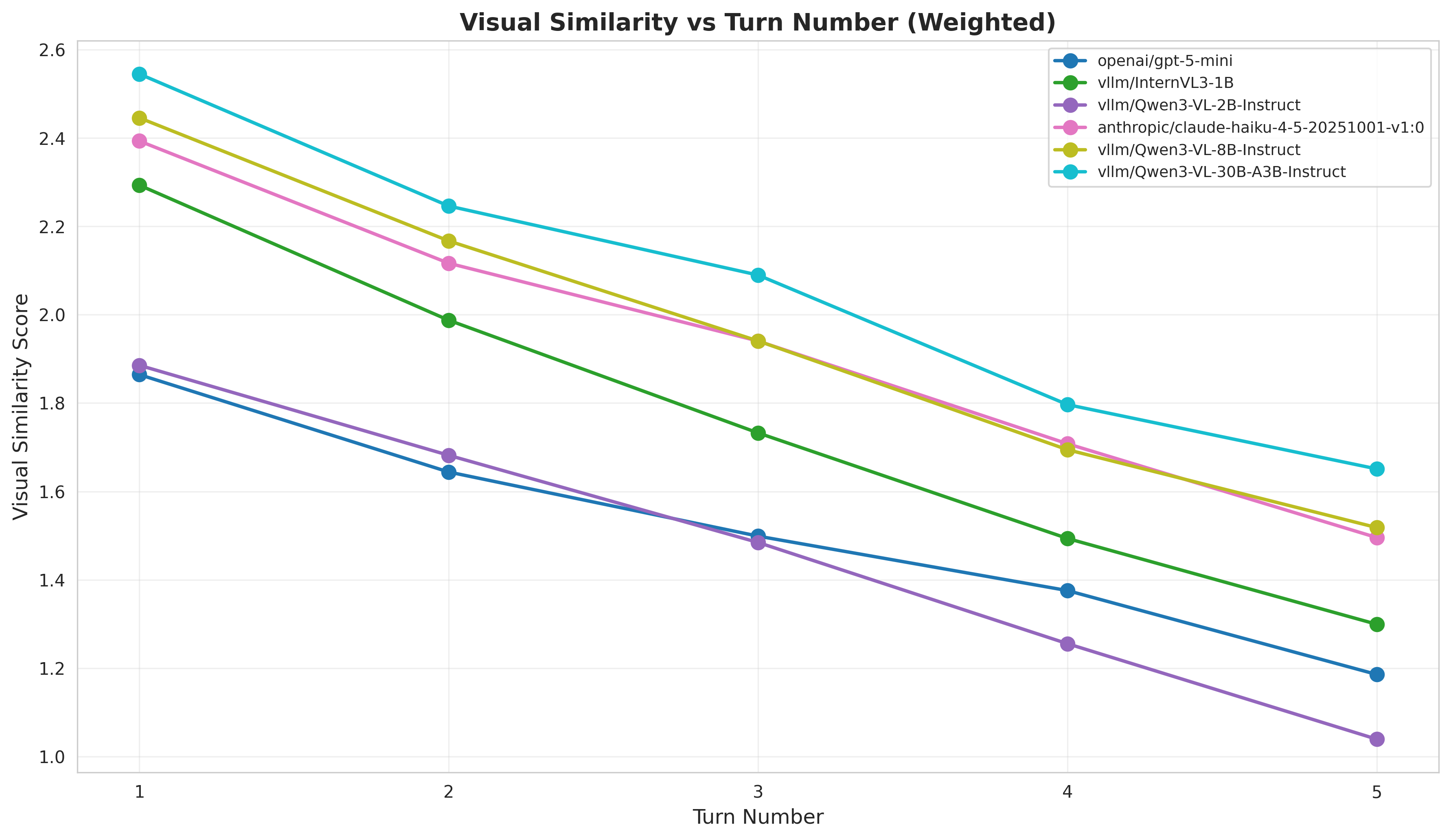}
\caption{Visual Similarity vs.\ Turn Number}
\label{fig:turn_vs}
\end{subfigure}
\hfill
\begin{subfigure}[t]{0.48\textwidth}
\centering
\includegraphics[width=\textwidth]{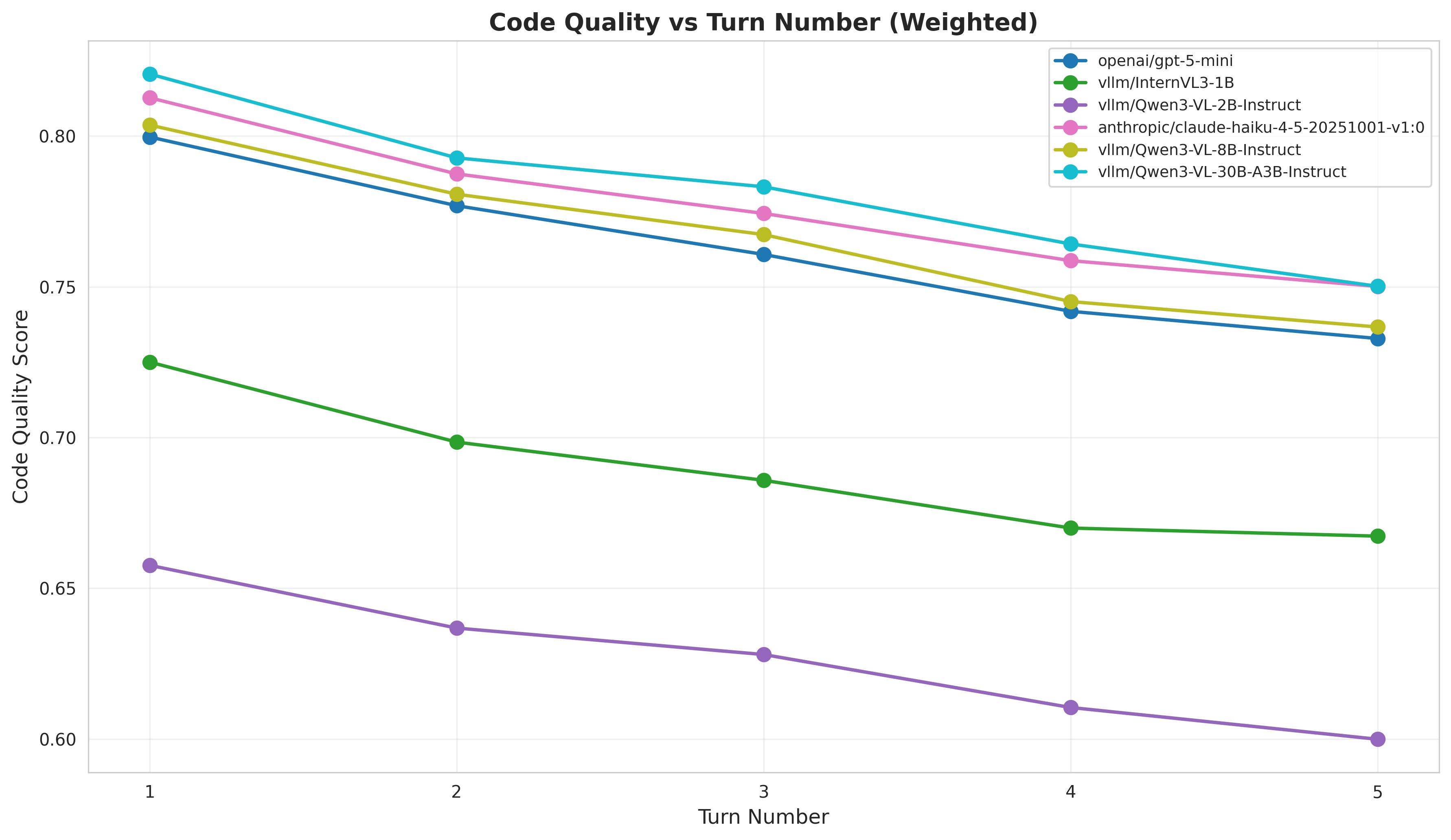}
\caption{Code Quality vs.\ Turn Number}
\label{fig:turn_cq}
\end{subfigure}
\caption{Detailed metric breakdown across conversation turns. Visual similarity (a) shows the steepest degradation, while code quality (b) remains more stable, indicating that accumulated errors primarily affect visual output rather than code structure.}
\label{fig:turn_metrics}
\end{figure*}

Figure~\ref{fig:turn_metrics} provides detailed per-metric analysis across conversation turns. Visual similarity (Figure~\ref{fig:turn_vs}) exhibits the most pronounced degradation, with Qwen3-VL-30B-A3B dropping from 2.55 to 1.65 (35\% decline) and GPT-5-mini from 1.87 to 1.19 (36\% decline). This steep decline occurs because visual similarity is computed against ground-truth charts that incorporate all previous modifications---any accumulated error in the modification chain compounds in the final visual comparison.

Code quality (Figure~\ref{fig:turn_cq}) shows more gradual decline, with most models dropping 5--10\% from turn 1 to turn 5. The relative stability of code quality metrics suggests that models maintain syntactically correct and well-structured code even when the semantic content deviates from ground truth. Claude Haiku 4.5 and Qwen3-VL-30B-A3B maintain the highest code quality throughout (approximately 0.75--0.81), while Qwen3-VL-2B shows the steepest decline (0.66 to 0.60).

\begin{figure}[h]
\centering
\includegraphics[width=\columnwidth]{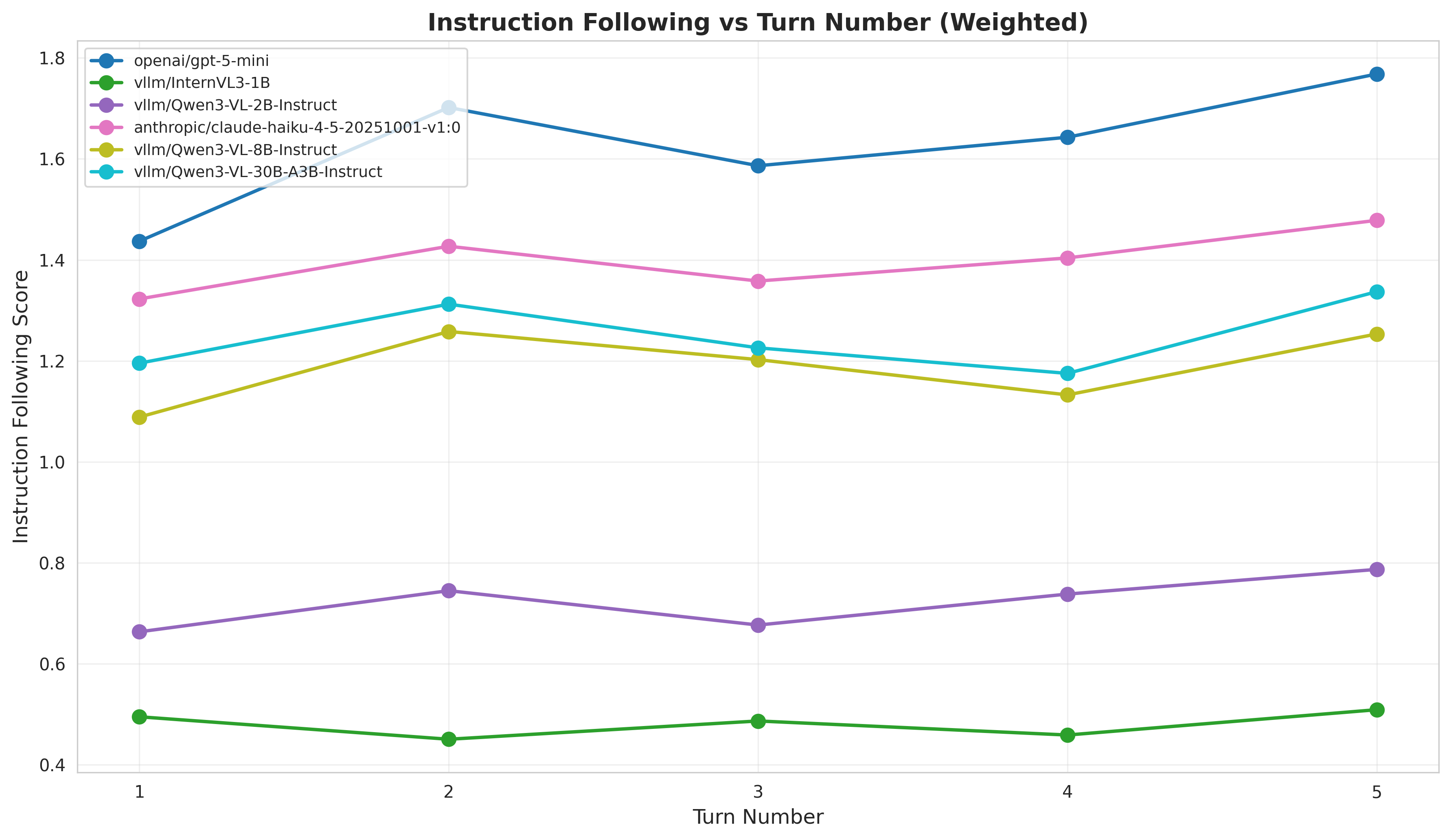}
\caption{Instruction following scores across turns (weighted). Scores remain stable or slightly increase, as instruction understanding is independent of accumulated rendering state.}
\label{fig:turn_if}
\end{figure}

Figure~\ref{fig:turn_if} shows instruction following scores across turns. Unlike other metrics, instruction following remains relatively flat or even slightly increases for some models. This counterintuitive pattern occurs because our fallback mechanism ensures models always receive valid input---when code fails to render, we use the last successful output. Thus, instruction following measures the model's ability to understand the current instruction independent of accumulated state, which remains stable across the conversation.

\subsection{Difficulty-Stratified Per-Metric Analysis}
\label{app:difficulty_metrics}

\begin{figure*}[t]
\centering
\begin{subfigure}[t]{0.32\textwidth}
\centering
\includegraphics[width=\textwidth]{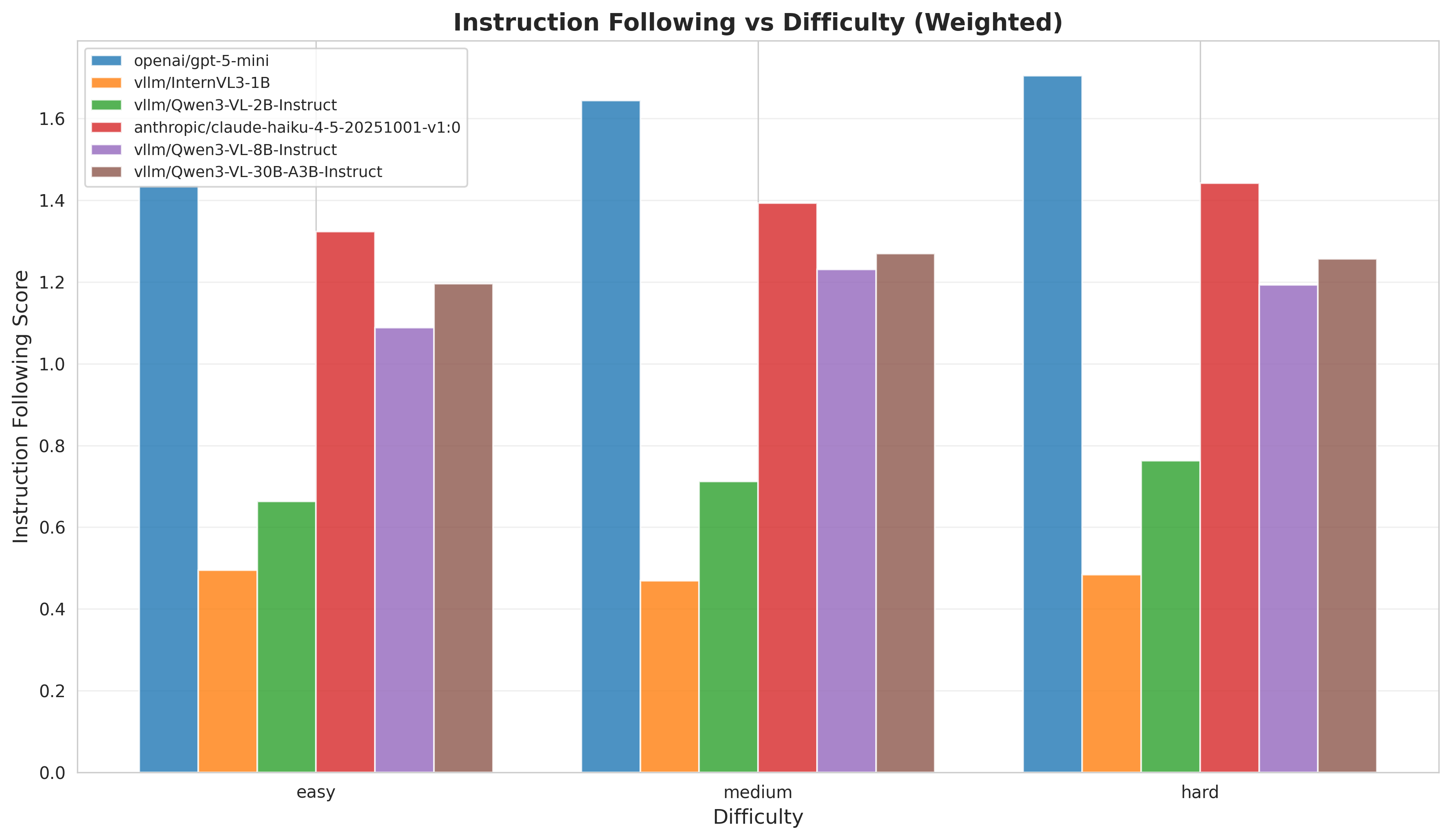}
\caption{Instruction Following}
\label{fig:difficulty_if}
\end{subfigure}
\hfill
\begin{subfigure}[t]{0.32\textwidth}
\centering
\includegraphics[width=\textwidth]{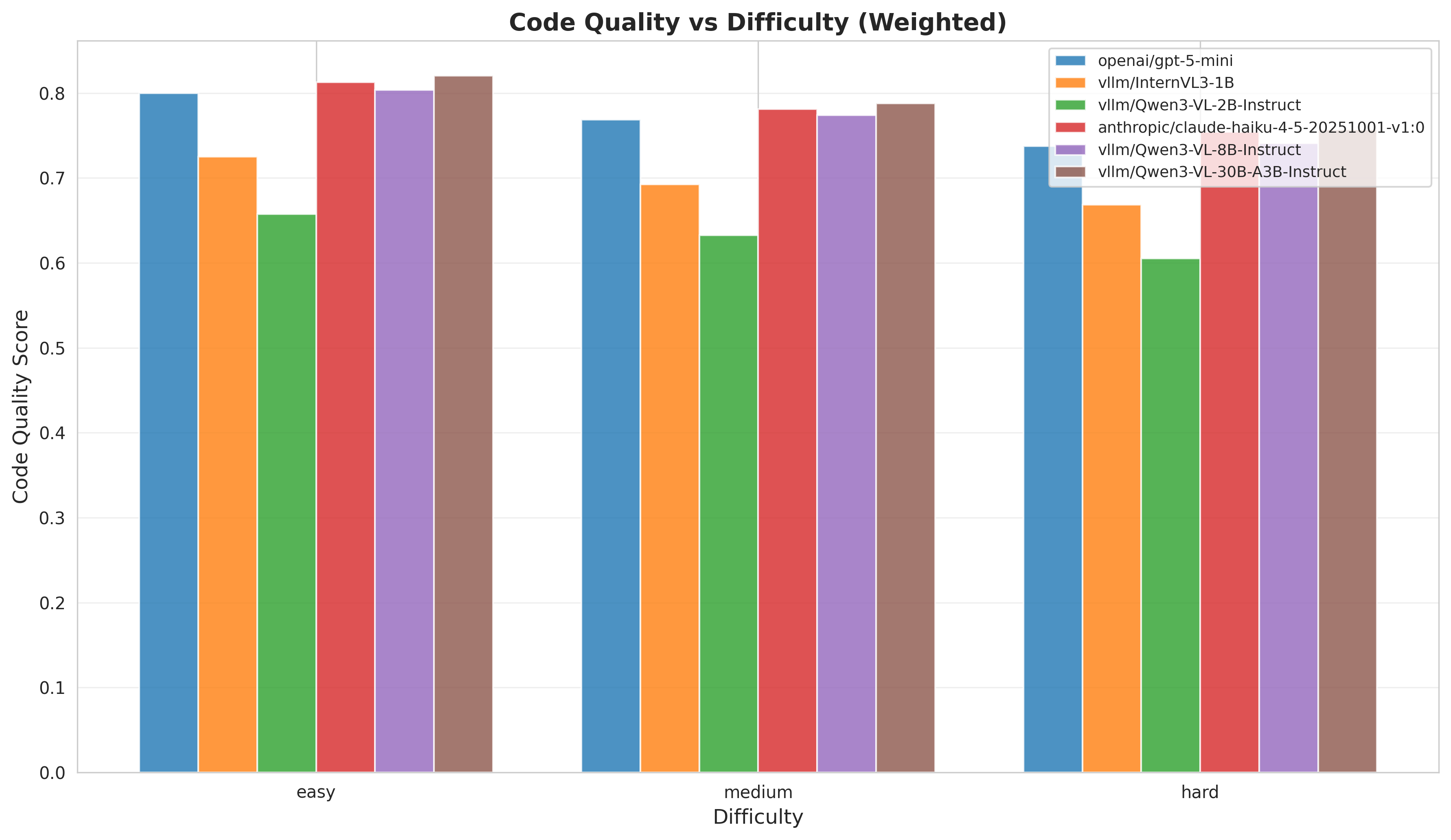}
\caption{Code Quality}
\label{fig:difficulty_cq}
\end{subfigure}
\hfill
\begin{subfigure}[t]{0.32\textwidth}
\centering
\includegraphics[width=\textwidth]{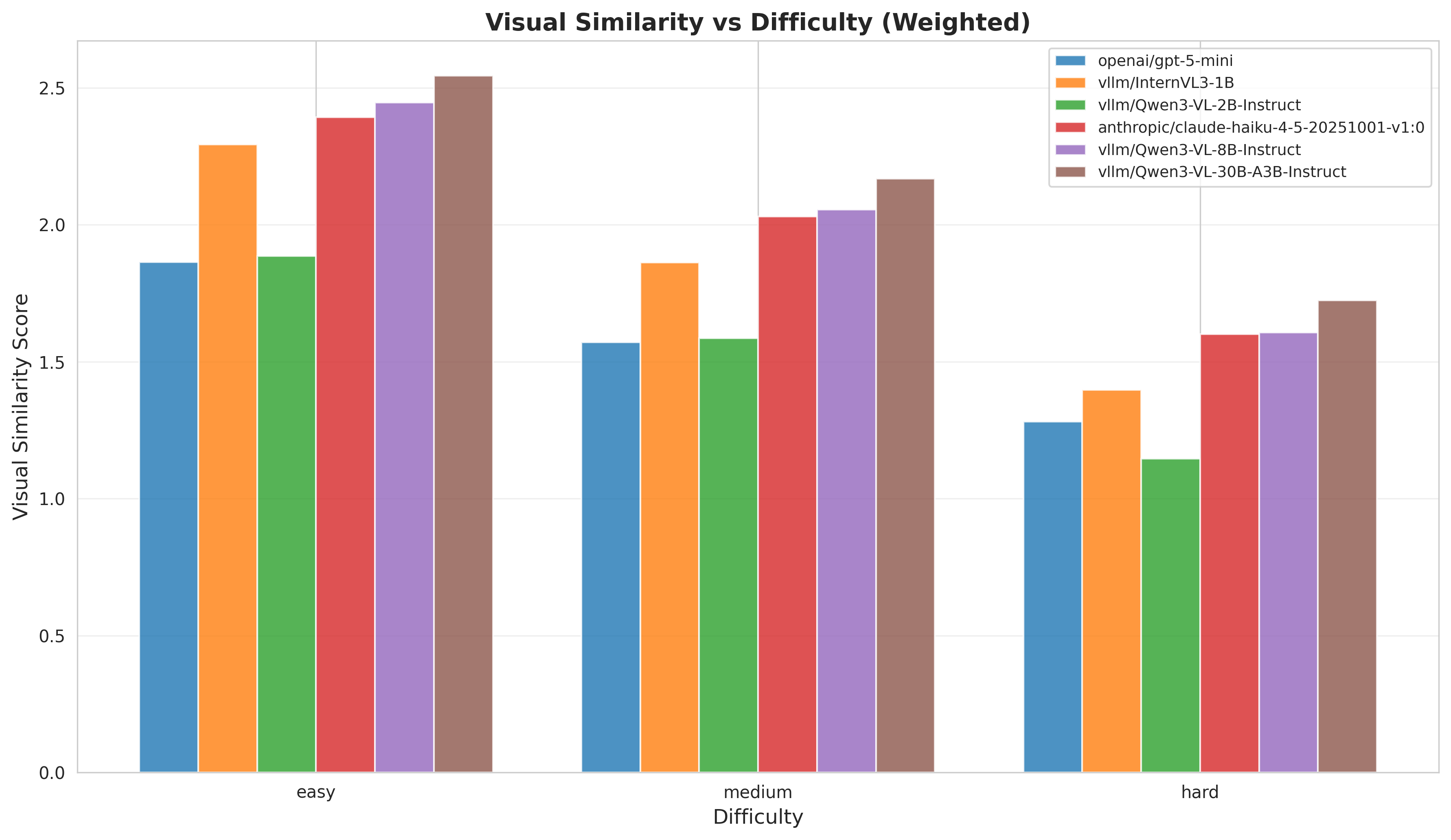}
\caption{Visual Similarity}
\label{fig:difficulty_vs}
\end{subfigure}
\caption{Per-metric performance breakdown across difficulty levels. (a) Instruction following shows unexpected increases for top models at higher difficulty, explained by the shift toward LLM-judged instructions. (b) Code quality remains stable across difficulties. (c) Visual similarity shows consistent degradation patterns.}
\label{fig:difficulty_metrics}
\end{figure*}

Figure~\ref{fig:difficulty_metrics} presents disaggregated analysis by metric across difficulty levels.

\paragraph{Instruction Following (Figure~\ref{fig:difficulty_if}).} A notable finding is that GPT-5-mini and Claude Haiku 4.5 show \textit{increasing} instruction following scores from easy (1.45, 1.32) to hard (1.70, 1.44). This counterintuitive result arises from our hybrid evaluation design: hard modifications more frequently involve subjective, semantic instructions evaluated by LLM judges, where top models excel. In contrast, smaller models (InternVL3-1B, Qwen3-VL-2B) show flat or slightly declining curves, indicating they struggle equally with both programmatic and semantic instructions.

\paragraph{Code Quality (Figure~\ref{fig:difficulty_cq}).} Code quality shows remarkable stability across difficulty levels for all models, with most variations within 0.05 points. This stability suggests that code quality assertions (import checking, structural validation, memory management) are largely orthogonal to modification difficulty. The Qwen3-VL models maintain the highest code quality (0.77--0.82), potentially reflecting training data that emphasized clean code practices.

\paragraph{Visual Similarity (Figure~\ref{fig:difficulty_vs}).} Visual similarity shows the expected monotonic decrease with difficulty. Qwen3-VL-30B-A3B achieves the highest easy-level score (2.55) but also shows the steepest absolute decline (to 1.72 for hard). GPT-5-mini consistently underperforms other top models on visual similarity, scoring 1.87 on easy modifications compared to 2.30--2.55 for others. This suggests GPT-5-mini may prioritize explicit instruction compliance over implicit visual consistency.

\subsection{Modification Type × Difficulty Interaction}
\label{app:modification_difficulty}

\begin{figure*}[t]
\centering
\includegraphics[width=\textwidth]{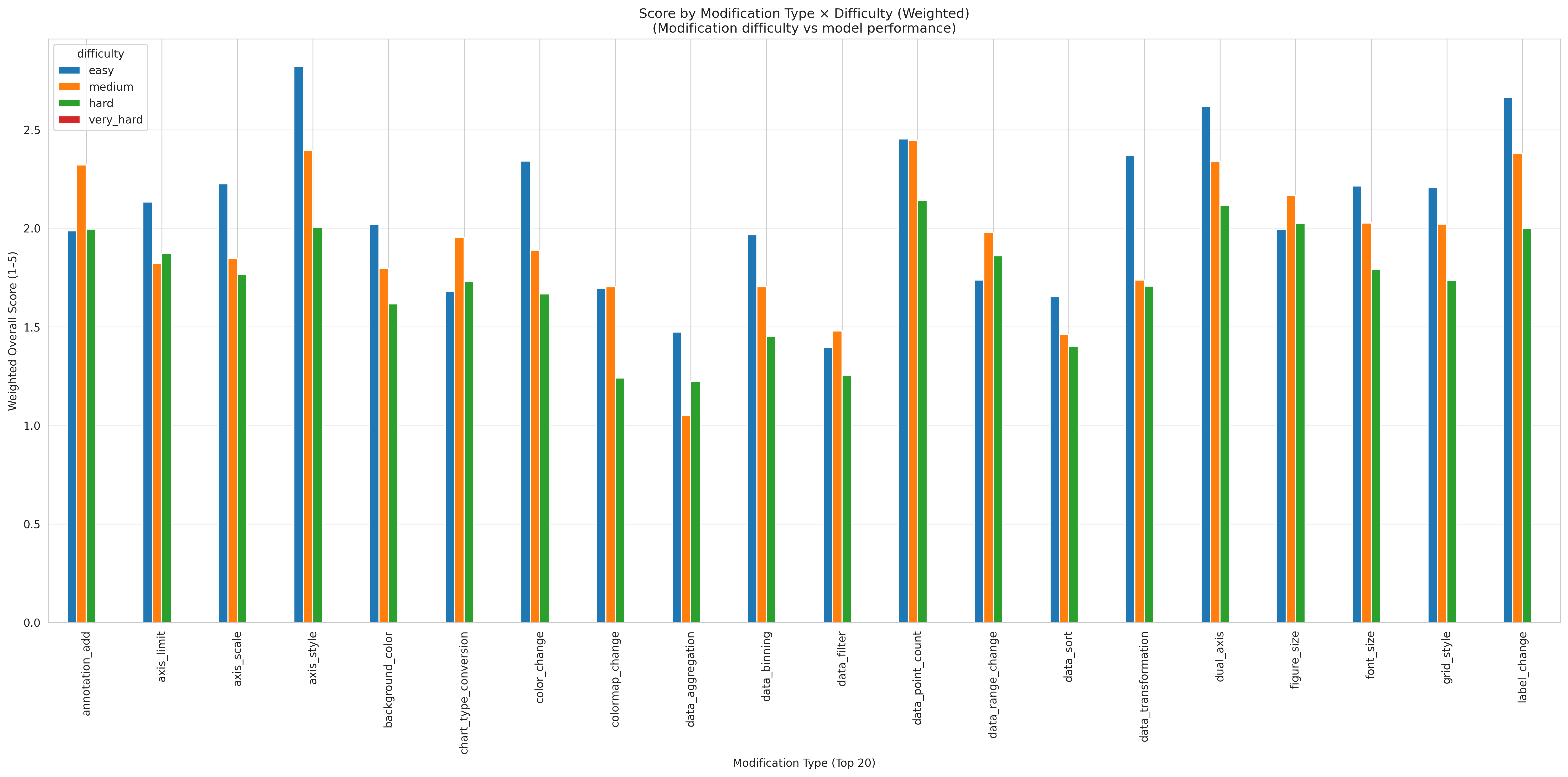}
\caption{Performance by modification type across difficulty levels (averaged over all models). Some modifications like \texttt{data\_aggregation} and \texttt{data\_filter} show low performance even at easy levels, while others like \texttt{grid\_style} maintain high performance across all difficulties.}
\label{fig:mod_difficulty}
\end{figure*}

Figure~\ref{fig:mod_difficulty} shows performance across modification types stratified by difficulty level. This reveals which modifications are inherently challenging versus those that become difficult only at higher complexity levels.

Several patterns emerge from this interaction analysis:

\begin{itemize}
    \item \textbf{Consistently Easy:} \texttt{axis\_style}, \texttt{grid\_style}, and \texttt{label\_change} maintain high scores ($>$2.0) even at hard difficulty, suggesting these modifications require similar skills regardless of complexity specification.
    
    \item \textbf{Consistently Difficult:} \texttt{data\_aggregation} and \texttt{data\_filter} score below 1.5 even at easy difficulty, indicating fundamental challenges in data manipulation that persist regardless of task complexity.
    
    \item \textbf{Difficulty-Sensitive:} \texttt{dual\_axis} and \texttt{data\_point\_count} show large drops from easy ($>$2.4) to hard ($<$2.1), suggesting that increased specification complexity substantially impacts performance.
\end{itemize}

\subsection{Rendering Success by Modification Type}
\label{app:rendering_success}

\begin{figure}[h]
\centering
\includegraphics[width=\columnwidth]{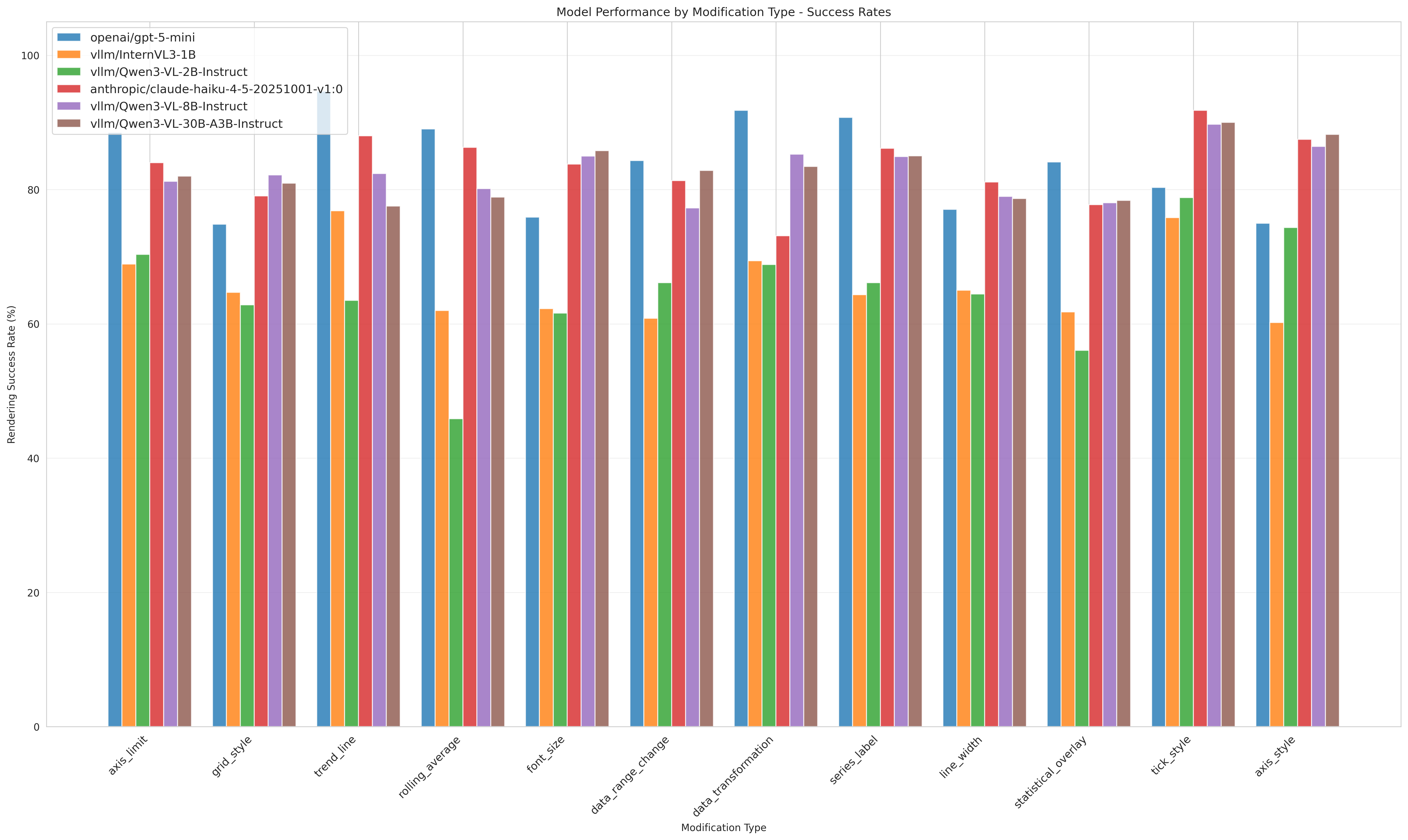}
\caption{Rendering success rates by modification type and model. Data manipulation operations (\texttt{rolling\_average}, \texttt{statistical\_overlay}) show lower success rates, particularly for smaller models.}
\label{fig:modification_success}
\end{figure}

Figure~\ref{fig:modification_success} presents rendering success rates---the percentage of model outputs that produce valid, executable code---across modification types. Key observations include:

\begin{itemize}
    \item \textbf{Trend line} modifications achieve near-perfect rendering success (90--100\%) for all models, as these involve simple additions to existing plot structures.
    
    \item \textbf{Rolling average} shows the lowest success rates, with Qwen3-VL-2B achieving only 46\% and InternVL3-1B at 62\%. These modifications require correct data indexing and window calculations that frequently cause runtime errors.
    
    \item \textbf{Statistical overlay} similarly challenges smaller models (55--65\% success), as computing and displaying statistical measures (mean lines, confidence intervals) requires precise data handling.
    
    \item \textbf{Proprietary models} (GPT-5-mini, Claude Haiku 4.5) maintain consistently high success rates (75--90\%) across all modification types, demonstrating robust code generation capabilities.
\end{itemize}

\subsection{Low Score Analysis}
\label{app:low_scores}

To understand failure modes, we analyze the rate of ``low scores'' (overall score $<$ 3.0 on the 5-point scale) by modification type and model.

\begin{figure}[h]
\centering
\includegraphics[width=\columnwidth]{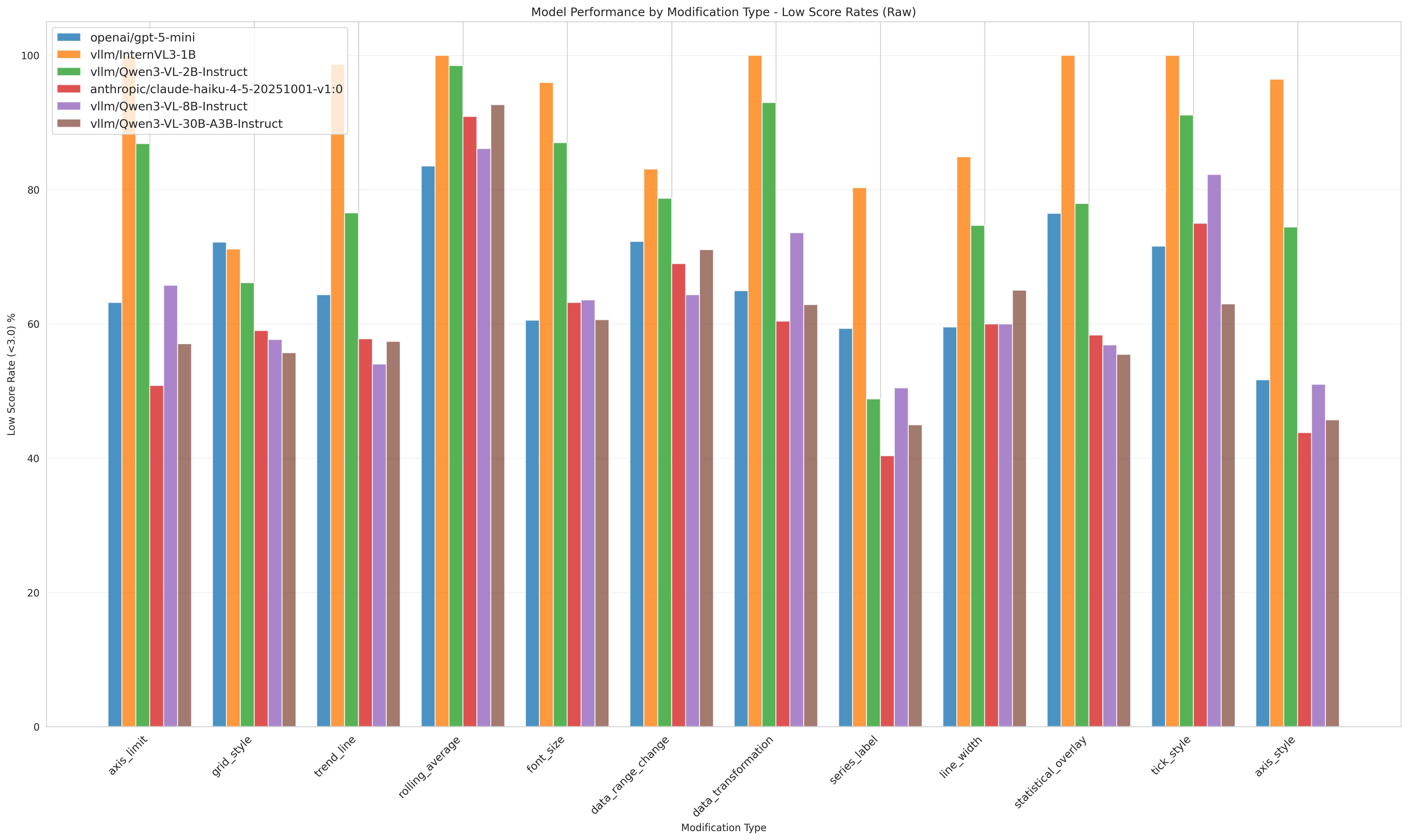}
\caption{Low score rates (score $<$ 3.0) by modification type and model. Higher bars indicate more frequent failures. Rolling average and data transformation show universally high failure rates.}
\label{fig:low_scores}
\end{figure}

Figure~\ref{fig:low_scores} reveals systematic failure patterns:

\begin{itemize}
    \item \textbf{Universal failures:} \texttt{rolling\_average} exceeds 80\% low-score rate for all models except Claude Haiku 4.5 (58\%), indicating this modification type is fundamentally challenging for current LMMs.
    
    \item \textbf{Model-specific patterns:} InternVL3-1B shows $>$85\% low-score rate on 8 of 12 modification types, while Claude Haiku 4.5 stays below 70\% on all types, demonstrating consistent capability advantages.
    
    \item \textbf{Relative strengths:} Claude Haiku 4.5 shows notably lower failure rates on \texttt{axis\_limit} (51\%) and \texttt{series\_label} (40\%) compared to other models (65--90\%), suggesting particular strength in label and axis manipulation.
\end{itemize}

\subsection{Overall Score by Difficulty (Bar Chart)}
\label{app:overall_difficulty_bar}

\begin{figure}[h]
\centering
\includegraphics[width=\columnwidth]{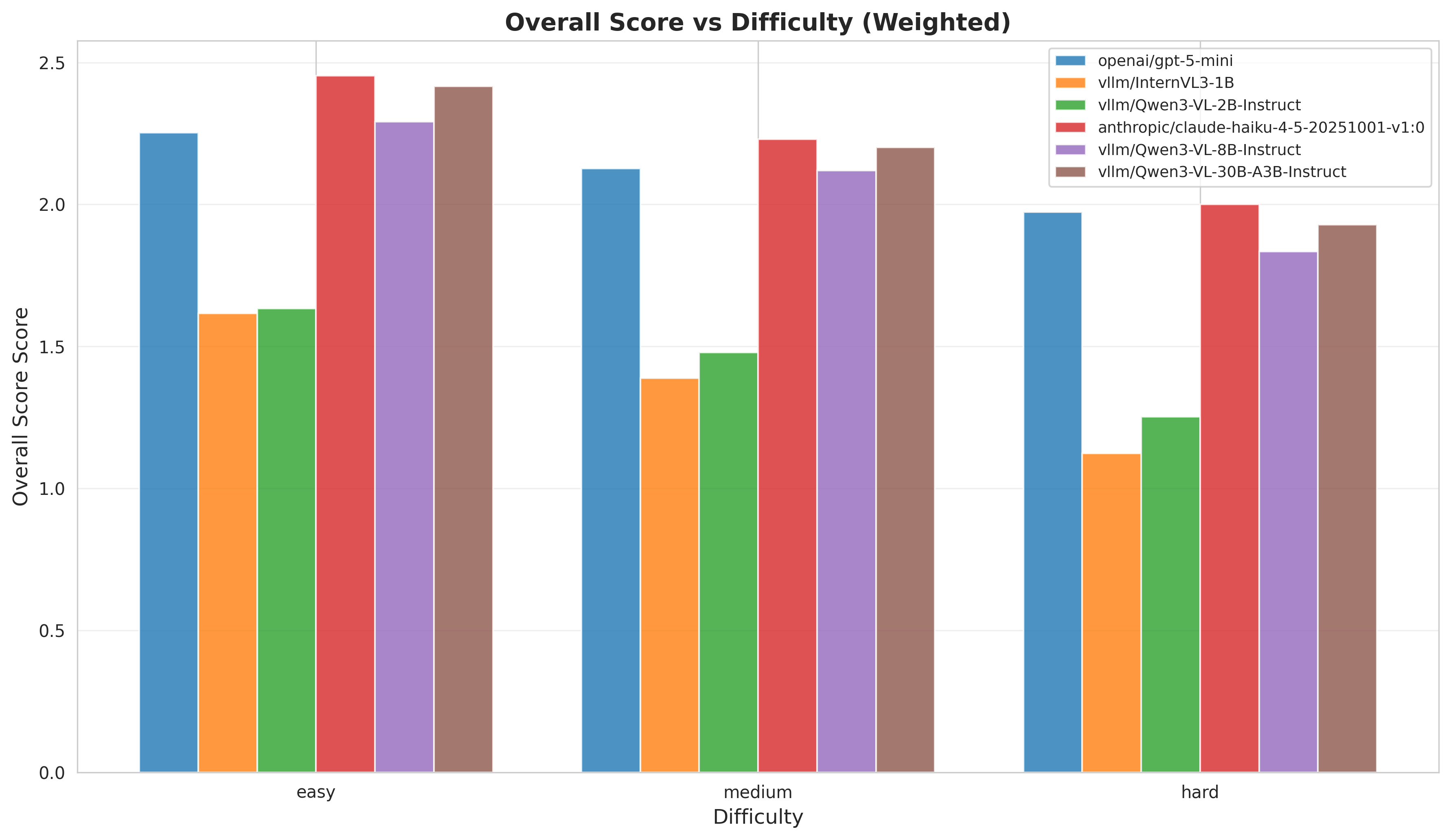}
\caption{Overall score by difficulty level (bar chart view). Provides alternative visualization of difficulty-stratified performance showing absolute score values.}
\label{fig:difficulty_overall_bar}
\end{figure}

Figure~\ref{fig:difficulty_overall_bar} provides a bar chart view of overall scores by difficulty, complementing the line chart in the main paper. This visualization emphasizes the absolute performance gap between model tiers: proprietary models and Qwen3-VL-30B maintain scores above 1.9 even on hard modifications, while smaller models drop below 1.3.

\section{Reproducibility Details}
\label{app:reproducibility}

\subsection{Random Seeds and Determinism}

All chart code generation uses fixed random seeds (\texttt{np.random.seed(42)}) to ensure identical data across experimental runs. Model inference uses temperature 0.1 for near-deterministic outputs while maintaining slight diversity for edge cases. vLLM serves open-source models with eager execution mode to ensure reproducible memory allocation patterns.

\subsection{Checkpoint and Logging}

The benchmarking system implements automatic checkpointing with resume capability. Each prediction is logged with:
\begin{itemize}
    \item Full model response text
    \item Extracted Python code
    \item Execution stdout/stderr
    \item Rendering success/failure status
    \item Timing information
    \item Fallback usage indicators
\end{itemize}

Outputs are organized in structured directories where each instance contains subdirectories for different turns, storing original, ground-truth, and predicted charts alongside their corresponding code files.

\subsection{Evaluation Metrics Computation}

All metrics are computed as weighted means where weights correspond to rendering success (1 for successful renders, 0 for failures). This weighting ensures that models are not penalized in quality metrics for failed renders (which receive separate tracking), while maintaining meaningful quality comparisons on successfully rendered outputs.

Final scores are aggregated as:
\begin{equation}
    \text{Score}_{\text{weighted}} = \frac{\sum_{i=1}^{N} w_i \cdot s_i}{\sum_{i=1}^{N} w_i}
\end{equation}
where $w_i = 1$ if sample $i$ rendered successfully and $w_i = 0$ otherwise.



\end{document}